\documentclass[11pt]{article}
\usepackage[preprint]{acl}
\usepackage{times}

\usepackage[T1]{fontenc}
\usepackage[utf8]{inputenc}
\usepackage{microtype}
\usepackage{inconsolata} 

\usepackage{latexsym}
\usepackage{graphicx}
\usepackage{amsmath}
\usepackage{amssymb}
\usepackage{booktabs}
\usepackage{xcolor}
\usepackage[normalem]{ulem}
\usepackage{listings}
\usepackage{url}
\usepackage{algorithm}
\usepackage{algorithmic}
\usepackage{tikz}
\usetikzlibrary{arrows.meta,positioning}
\usepackage{lmodern}

\title{ScalableRAG: High-Quality RAG with Exact Corpus Operations at Minimal Ingestion Cost}

\author{
  Hilaf Hasson\textsuperscript{1},
  Aditya Chakravarty\textsuperscript{1},
  Jayant Thomas\textsuperscript{1},
  Krishna Gogineni\textsuperscript{1} \\[0.4em]
  \normalfont \textsuperscript{1}Cohesity \\[0.2em]
  \textsuperscript{1}{\fontfamily{lmtt}\selectfont
  \char`\{hilaf.hasson,aditya.chakravarty,jayant.thomas,krishna.gogineni\char`\}@cohesity.com}
}

\begin{document}
\maketitle

\begin{abstract}

RAG systems face a capability-cost gap: lightweight retrievers locate evidence but expose no operands for exact corpus-level operations (counts, intersections, aggregations), while knowledge ingestion systems obtain them only through expensive ingestion. \textbf{ScalableRAG} closes this gap with a persistent typed-set workspace: an agent builds named document and value sets at query time and composes them through set algebra, typed extraction, aggregation, and diagnostics.

\textbf{Zero-Ingestion ScalableRAG} needs no preprocessing, not even a vector index. It matches the leading baseline on four conventional QA benchmarks (74.75 versus 74.64) and on aggregative Hotels-50 beats \emph{every} evaluated system, scoring 71.74: 36.23 above the leading baseline (SRAG), and 40.22 points above A-RAG. Removing the workspace drops Hotels-50 to 38.4. On Hotels-350, the partition SRAG itself reports on, ScalableRAG leads SRAG by 9.3 points. \textbf{Limited-Ingestion ScalableRAG} adds a lightweight index and is the only evaluated system above 80 judge score at 5M documents, at 13 to 16 seconds per question and under 20 minutes ingestion; at 1M it scores 84.5 to A-RAG's 80.0 with 16$\times$ lower latency and $0.5\%$ of its ingestion time. ScalableRAG combines minimal ingestion, conventional-QA parity, exact corpus operations, and million-document viability; no evaluated alternative provides all four.

%We also introduce a lightweight indexing version: Limited-Ingestion
%ScalableRAG; which allows Zero-Ingestion ScalableRAG to scale to >1M doc corpora, where it is the only viable high performance retrieval system.

Our code is available at \href{https://github.com/cohesity/ScalableRAG}{https://github.com/cohesity/ScalableRAG}.
\end{abstract}

%=======================================================================
\section{Introduction}
\label{sec:intro}
%=======================================================================

RAG systems face a costly choice. Lightweight retrievers \citep{karpukhin2020dense, lewis2020retrieval} and agents can locate local evidence, but their intermediate results usually disappear into textual context. They therefore provide no complete operand for exhaustive filtering, counting, intersection, or aggregation. Knowledge-graph \citep{edge2024local, bai2025autoschemakg, gutierrez2024hipporag, gutierrez2025rag} and schema-first \citep{koshorek2025structured} systems materialize corpus structure offline, but pay substantial ingestion cost before answering a question. ScalableRAG removes this tradeoff: it materializes only the structure needed for the current question, preserves that structure as reusable sets, and executes deterministic corpus operations over it.

 %ScalableRAG's key abstraction is to augment the agentic retriever with a registry of multiple named document sets and typed value sets. %During retrieval, steps can refine, intersect, union, subtract, project from, and aggregate those relations without reconstructing earlier results in the LLM context. Each derived object retains provenance, and set-relative diagnostics expose how the new set differs from its parent. Thus ScalableRAG incrementally, at inference, constructs an executable retrieval program whose set registry state survives even as textual observations are compressed.

% The workspace is an abstract data type, not a cache or a memory add-on: its
% operators are defined over the named relations it stores. Set algebra needs
% reusable operands, reductions need complete value relations, and
% parent-relative diagnostics need stored lineage. ScalableRAG therefore records
% not only what text to read but what relations the question has constructed,
% and the relations survive after textual observations are compressed. The
% resulting operating point---minimal ingestion, ordinary-QA quality, exact
% corpus operations, and large-corpus access in a single architecture---is
% offered by none of the systems we evaluate, including the ingestion-heavy
% graph and schema architectures built for structured operations.

ScalableRAG's key abstraction is to augment the agentic retriever with a registry of multiple named document sets and typed value sets. \textbf{Zero-Ingestion ScalableRAG} requires no corpus preprocessing or vector
index. At inference, it creates and persists named document sets and typed sets of numbers,
dates, and lists; later steps combine and aggregate them while parent-relative
diagnostics expose implausible narrowing. Starting from raw documents, it not
only beats every evaluated baseline on the aggregative Hotels datasets, but
also matches the strongest agentic baseline on four conventional (i.e., not made up of aggregative questions) QA datasets, showing
 that adding exhaustive corpus operations does not trade
away ordinary retrieval quality even without an embedding index.

The zero-ingestion design uses regex for set creation, which is fast on the corpora in our main evaluation but requires repeated scans as the corpus grows. \textbf{Limited-Ingestion ScalableRAG} makes the same workspace scale: a lightweight index cheaply materializes complete named candidate sets from literal queries when the parent set is large, and later operations reuse and refine those sets instead of rescanning the entire corpus. Full regex filtering returns once a set is small. Limited-Ingestion is the only evaluated system to complete at 5M with GPT-4.1 judge score above 80, maintaining 13--16 second mean inference latency with 1{,}166.01 seconds of ingestion (Table \ref{tab:scale}).

% \paragraph{Evaluation design.}
% The evaluation follows the architecture. The Hotels corpora and Transcripts test exact
% corpus operations; four established QA benchmarks test whether the same
% interface retains ordinary retrieval quality; a nested-corpus study scales the
% search space to 5M documents while holding the questions fixed; and a matched
% comparison that removes only the persistent workspace isolates the mechanism
% itself, dropping the aggregative Hotels-50 score from 71.74 to 38.4. The
% boundary between large-corpus access and downstream operations is an exact
% named relation: once registered, the same operators execute independently of
% documents outside that relation, so the capability and scale studies compose at
% the system's explicit interface. Throughout, exact operations pay their
% unavoidable output-sensitive cost only for question-relevant relations;
% the ambient corpus never requires graph or schema materialization.

% Together, these two solutions offer a practical middle ground between heavy ingestion (one or more LLM calls per document) and stateless retrieval agents: ScalableRAG achieves strong end-to-end accuracy while preserving a rich set-algebraic tool surface at inference time, and keeping preprocessing costs minimal.

Our contributions are:
\begin{enumerate}
   \item \textbf{The persistent typed-set workspace}: a closed interface of named document and value sets with set algebra, typed extraction, deterministic aggregation, provenance, and automatic set-relative diagnostics (Section \ref{ssec:zero}), executed in two regimes sharing one state abstraction: \textbf{Zero-Ingestion ScalableRAG}, which begins from raw documents with full-scan filtering and no index of any kind, and \textbf{Limited-Ingestion ScalableRAG}, which adds lightweight indexing for million-document corpora (Section \ref{ssec:limited}).
   % \item \textbf{Transcripts}, a new 100-question benchmark over 75 Congressional hearing transcripts that mixes local lookups with corpus-level operations; all golds are generated deterministically and recomputed from scratch by an independent validator, enabling exact evaluation of corpus-level reasoning (Appendix~\ref{app:chrg}).
   \item \textbf{A new quality--ingestion frontier}, established on six public benchmarks and Transcripts, a new corpus-operations benchmark with independently validated golds (Appendix~\ref{app:chrg}): Zero-Ingestion matches the strongest agentic baseline on conventional QA while beating every evaluated system (agentic, graph, and schema) on aggregative QA by 36.23 points over the best baseline on Hotels-50, and remains the strongest system on Hotels-350, the 350-document partition on which SRAG's authors, who introduced the dataset, report their results. Further, Limited-Ingestion is the only system completing 5M documents within a 48-hour budget with judge score above 80, and at 1M it matches A-RAG's quality (84.5 versus 80.0) with 16$\times$ lower latency and $0.5\%$ of its ingestion time. A matched ablation attributes the aggregative capability to the workspace itself (Section \ref{sec:experiments}).
    %vector embeddings, and a thorough pattern and extraction discovery and validation pre-processing so as to improve performance with a \emph{constant} number of LLM calls. This leads to moderate accuracy improvements in those datasets where extraction is more challenging.
\end{enumerate}
%=======================================================================
\section{Related Work}
\label{sec:related}
%=======================================================================
\paragraph{Baseline RAG.}
Dense RAG returns ranked passages as context~\citep{karpukhin2020dense, lewis2020retrieval}. HyDE and HyQE improve query/document representations~\citep{gao2023precise,zhou2024hyqe}, while FLARE, Iter-RetGen, and IRCoT make retrieval iterative~\citep{jiang2023active,shao2023enhancing,trivedi2023interleaving}; their retrieved passages nevertheless remain textual context rather than executable operations.

\paragraph{Ingestion-Heavy Knowledge Representation}
GraphRAG \citep{edge2024local}, HippoRAG \citep{gutierrez2024hipporag}, and HippoRAG2 \citep{gutierrez2025rag} materialize entity-relation graphs and graph indexes; AutoSchemaKG~\citep{bai2025autoschemakg,peng2025graph} induces schemas and builds a triple index at web scale. SRAG~\citep{koshorek2025structured} instead populates a one-row-per-document SQL table through LLM extraction. These systems gain structured operations by paying substantial offline graph construction, indexing, or per-document extraction cost. ScalableRAG materializes document-keyed structure only as questions require it.

\begin{figure*}[t]
\centering
\definecolor{srblue}{HTML}{2A5D8F}
\definecolor{srbluefill}{HTML}{EAF2F8}
\definecolor{srgrayfill}{HTML}{F4F5F6}
\resizebox{0.96\textwidth}{!}{%
\begin{tikzpicture}[
  x=1cm,y=1cm,
  font=\sffamily\footnotesize,
  box/.style={draw=black!55, rounded corners=2pt, align=center,
              inner xsep=8pt, inner ysep=7pt, line width=.55pt},
  endpoint/.style={box, fill=white, minimum width=1.85cm},
  component/.style={box, fill=srgrayfill, minimum height=1.35cm},
  state/.style={box, draw=srblue, fill=srbluefill,
                minimum height=1.55cm, line width=.8pt},
  flow/.style={-{Latex[length=2.2mm]}, line width=.7pt, draw=black!68},
  stateflow/.style={-{Latex[length=2.2mm]}, line width=.75pt, draw=srblue}
]
  \node[endpoint] (question) at (0,0) {Question $q$};
  \node[component, text width=2.25cm] (agent) at (3.10,0)
        {\textbf{LLM agent}\\plan next step};
  \node[component, text width=3.05cm] (tools) at (7.35,0)
        {\textbf{Corpus tools}\\filter $\cdot$ read \\
          $\cdot$ set operations $\cdot$ aggregate};
  \node[state, text width=4.85cm] (workspace) at (12.45,0)
        {\textbf{Workspace of persistent sets}\\
         document sets $\mathcal{S}^{(t)}$\\
         (initialized with $\texttt{all}=\mathcal{C}$)\\
         typed value sets $\mathcal{V}^{(t)}$, provenance, and\\
         pointers to corpus documents};
  \node[endpoint] (answer) at (3.10,1.75) {Answer};

  \draw[flow] (question) -- (agent);
  \draw[flow] (agent.north) -- node[right, font=\sffamily\scriptsize] {final}
        (answer.south);

  \draw[flow] (agent.north east) to[out=35,in=145]
        node[above, font=\sffamily\scriptsize, yshift=0pt]
        {tool + named inputs} (tools.north west);
  \draw[stateflow] (tools.south west) to[out=215,in=325]
        node[below, font=\sffamily\scriptsize, yshift=-2pt]
        {tool observation} (agent.south east);

  \draw[stateflow] (tools.north east) to[out=35,in=145]
        node[above, font=\sffamily\scriptsize, yshift=5pt]
        {set creation} (workspace.north west);
  \draw[stateflow] (workspace.south west) to[out=215,in=325]
        node[below, font=\sffamily\scriptsize, yshift=-2pt]
        {set details and diagnostics}
        (tools.south east);
\end{tikzpicture}
}%
\caption{ScalableRAG inference loop.  The agent invokes corpus tools using
named sets.  The tools obtain set details, diagnostics, and document pointers
from the workspace, and create new persistent document or value sets.  Tool
observations guide subsequent steps until the agent returns an answer.}
\label{fig:overview}
\end{figure*}

\paragraph{Agentic RAG.}
%Iter-RetGen~\citep{shao2023enhancing} iterates retrieval and generation.
%IRCoT~\citep{trivedi2023ircot} interleaves retrieval with chain-of-thought.
%Recent work pushes further toward genuine agents.
A-RAG~\citep{du2026rag} is an agent equipped with three retrieval tools:
\texttt{keyword\_search} for exact lexical matching that returns snippet-level evidence,
\texttt{semantic\_search} for dense sentence-level retrieval (using a vector database) grouped by chunks, and
\texttt{chunk\_read} for selected full text. A-RAG provides our primary agentic contrast: it is a strong iterative retriever, while its intermediate results remain textual observations rather than operands in an external set registry.

% \paragraph{Semantic operators and code agents.}
% LOTUS~\citep{patel2024lotus}, DocETL~\citep{shankar2024docetl}, and Palimpzest~\citep{liu2024palimpzest} run declarative filter, map, and aggregate pipelines over document collections by calling an LLM per document per operator, so their cost grows with the corpus as ingestion does; ScalableRAG builds sets with LLM-free predicates and spends LLM calls only on planning. Program-aided agents~\citep{gao2023pal} can implement set algebra ad hoc in a code interpreter; the workspace differs in exposing set state as a typed, named interface with automatic set-relative diagnostics.

\section{Proposed Method}
\label{sec:method}

We describe ScalableRAG in two stages.  Section~\ref{ssec:zero} presents the
\emph{zero-ingestion} system: a stateful agent whose sole input is a directory
of plain-text files.  Section~\ref{ssec:limited} retains the same persistent
state abstraction but introduces a size-aware access contract based on indexed
narrowing, cached patterns, and optional coarse embeddings.  Figure~\ref{fig:overview}
shows how the agent, corpus tools, and persistent-set workspace interact during
inference.

\subsection{Zero-Ingestion ScalableRAG}
\label{ssec:zero}
%=======================================================================

The zero-ingestion system requires no preprocessing: not only no knowledge graph or schema extraction, but also no vector embeddings.  The agent is given only the corpus and a question.  

\subsubsection{Agent Architecture}
\label{sssec:arch}

Let $\mathcal{C} = \{d_1, \dots, d_N\}$ be the corpus of $N$ documents.

ScalableRAG is at its base a ReAct-style agent~\citep{yao2022react} that interacts with the corpus through tool calls.  Given a question $q$, the agent produces a sequence of $(a_t, o_t)$ pairs, where $a_t$ is a JSON-formatted tool invocation and $o_t$ is the tool observation, until it emits a final answer.  

However, unlike a ReAct-style agent, at turn $t$, the agent maintains a \emph{set registry}~$\mathcal{S}^{(t)} = \{(s_i, D_i)\}_i$, where for each $i$ the set $D_i$ is a subset of $\mathcal{C}$, and its name is $s_i$.
The registry is initialized to $\mathcal{S}^{(0)} = \{(\texttt{all}, \mathcal{C})\}$ and is updated in-place after every tool invocation.
In addition, the agent maintains a set of \emph{value sets} $\mathcal{V}^{(t)} = \{(v_j, f_j)\}_j$, where $v_j$ names a partial map $f_j : D \rightharpoonup \mathbb{R}$ (scalar), $f_j : D \rightharpoonup \mathcal{T}$ (date), or $f_j : D \rightharpoonup \operatorname{Seq}(\Sigma^{*})$ (finite sequence of strings, preserving order and multiplicity), where $D$ is one of the sets in $\mathcal{S}^{(t)}$ and $\mathcal{T}$ is the date domain. Undefined entries are documents from which extraction produced no value; extraction coverage is reported to the agent.

At each step the agent selects a tool $\tau_t$ and arguments $\alpha_t$; the tool reads from and writes to the shared state:
\begin{equation}
\label{eq:step}
(\mathcal{S}^{(t)}, \mathcal{V}^{(t)}, o_t) = \tau_t(\alpha_t; \mathcal{S}^{(t-1)}, \mathcal{V}^{(t-1)}, \mathcal{C})
\end{equation}

This differs from stateless agentic RAG solutions, and allows the agent to gain significant helpful knowledge at each turn by automating computations that refer to existing sets; see Section \ref{sssec:diagnostics} for details.

\paragraph{What exactness means.}
Every registered set is complete under its declared predicate and every reduction is computed over the entire set; this is \emph{execution} exactness. Whether the predicate captures the natural-language condition is the agent's decision, which the diagnostics of Section~\ref{sssec:diagnostics} support but cannot certify. The aggregative benchmarks test this second property directly: each Transcripts and Hotels count or aggregate gold is a deterministic function of a gold document set, and the judge credits counts only on exact numerical match, so a system scores only when the set it constructed yields the same value as the gold set (Appendix~\ref{app:chrg}).

The full system prompt is given verbatim in Appendix~\ref{app:prompt}; and context management details (per-tool output budgets, observation redaction after consumption) are described in Appendix~\ref{app:context}.

\paragraph{Lexical pre-analysis.}
Before the first LLM turn, ScalableRAG performs an LLM-free scan to give the agent a coarse map of the search space.  Concretely, it extracts up to five distinct \emph{question keywords} (maximal alphanumeric spans of length $\geq 4$) and, for each keyword, counts how many documents contain it by case-insensitive substring inclusion.  It also reports filename hit counts, up to three nontrivial pairwise intersections, and one bounded seed snippet from the most selective keyword. Together these form a deterministic lexical first pass that helps the agent choose filenames, a broad text filter, or a narrower combination; the same procedure is used on every corpus.  
%The system also selects a small set of \emph{anchor tokens} (a heuristic subset of selective question keywords) that are reused only for diagnostics such as scope checks and for centering negative snippets.  
The question keywords above get sorted based on a lightweight logic (see Appendix~\ref{app:context}), and are henceforth called ``anchor tokens''.

Algorithm~\ref{alg:agent} formalizes the agent loop.

\begin{algorithm}[t]
\small
\caption{ScalableRAG Agent Loop}
\label{alg:agent}
\begin{algorithmic}[1]
\REQUIRE Corpus $\mathcal{C}$, question $q$, LLM $\mathcal{M}$, max steps $T$, context budget $B$
\STATE $\mathcal{S}^{(0)} \gets \{(\texttt{all}, \mathcal{C})\}$; \quad $\mathcal{V}^{(0)} \gets \emptyset$
\STATE $\ell \gets \textsc{QuestionLandscape}(q, \mathcal{C})$ \hfill \COMMENT{keyword statistics}
\STATE $\text{msgs} \gets [\text{SysPrompt}, \; q \oplus \ell]$
\FOR{$t = 1, \ldots, T$}
  \STATE $\text{msgs} \gets \textsc{RedactAndCompress}(\text{msgs}, B)$ \hfill \COMMENT{context management}
  % \STATE $\text{msgs} \gets \textsc{Compress}(\text{msgs}, B)$ \hfill \COMMENT{fit context budget}
  \STATE $\text{response} \gets \mathcal{M}(\text{msgs})$
  \IF{response contains \texttt{answer}}
    \RETURN response.\texttt{answer}
  \ENDIF
  \STATE Parse $(\tau_t, \alpha_t) \gets$ response.\texttt{tool}, response.\texttt{args}
  \STATE $(\mathcal{S}^{(t)}, \mathcal{V}^{(t)}, o_t) \gets \tau_t(\alpha_t; \mathcal{S}^{(t-1)}, \mathcal{V}^{(t-1)}, \mathcal{C})$
  \STATE Append assistant response and $o_t$ to msgs
\ENDFOR
\STATE \RETURN $\mathcal{M}(\text{msgs} \oplus \text{``give your best answer''})$
\end{algorithmic}
\end{algorithm}

% \subsubsection{The Typed Persistent Workspace}
% \label{sssec:tools}

% The central design choice of ScalableRAG is that every tool operates on, and writes back to, a shared, typed, named \emph{workspace} that persists across the entire reasoning trajectory.  This is what distinguishes our approach from all prior agentic RAG systems, where each tool call is a stateless query that returns text into the conversation and leaves no reusable artifact behind.

% Let $\mathcal{C}$ be the corpus. The workspace contains two kinds of named objects:
% \begin{enumerate}\setlength{\itemsep}{0pt}
% \item \textbf{Document sets} $\mathcal{S} = \{(s_i, D_i)\}_i$: subsets $D_i \subseteq \mathcal{C}$ with associated names $s_i$.
% \item \textbf{Value sets} $\mathcal{V} = \{(v_j, f_j)\}_j$: named per-document mappings $f_j : D \to T$, where $D\in \mathcal{S}$ with associated names $v_j$, and where $T$ is $\mathbb{R}$ (scalars), text, dates, or $2^{\Sigma^*}$ (lists of strings).
% \end{enumerate}

% Every tool \emph{consumes} named objects from the workspace and \emph{produces} new named objects back into it.  The agent refers to these objects by name in subsequent calls, composing multi-step programs.  

\paragraph{Producing document sets.}
Every filtering operation targets an existing named set rather than the corpus
implicitly.  The primary operation,
$\texttt{apply\_filter}(\rho,S)\rightarrow(S^+,S^-)$, partitions $S$ into the set of
documents $S^+$ that match $\rho$ and their complement $S^-$, and registers both sets.
Filename search, set intersection/union/difference, and predicates over
previously extracted values likewise produce named document sets.  This
relative semantics is essential: each turn refines the current working set,
and its evidence and diagnostics are scoped to that set.

\paragraph{Producing value sets.}
Extraction lifts a document set into a typed per-document mapping using agent-generated regex followed by a type conversion (into values/dates/etc.).  A capture
may store one scalar, date, or string per document, or a list of repeated
values; filenames can be treated as another metadata source. %Failed or
%missing extractions remain absent rather than being imputed.

\paragraph{Reducing to answers.}
Value sets support scalar and grouped reductions, including counts, sums,
averages, extrema, unique values, and value frequencies.  An extremum can also
return the corresponding documents for inspection; document-set cardinality
and arithmetic are evaluated deterministically.

\paragraph{Observation and planning.}
The agent can inspect sampled documents, page through long documents, navigate
to exact or query-relevant offsets, probe tables, compare candidate regexes,
and inspect set provenance.  For deterministic tools, it may preview several
automatically generated alternatives and select one; only the selected branch
is committed.  Appendix~\ref{app:prompt} gives the complete tool interface.
\subsubsection{Diagnostic Feedback: Using Set Persistence to Compute Intermediate Validation Hints}
\label{sssec:diagnostics}

Set persistence also enables automatically computed diagnostics that are unavailable to a stateless retrieval interface. We enumerate these for the primary tools.

\paragraph{\texttt{apply\_filter} diagnostics.}
When the agent partitions a target set $S$ by regex $\rho$, the observation contains far more than just the resulting set names and sizes.  All statistics below are computed \emph{relative to $S$}, so when the agent filters a $m$ document subset, it sees token hit counts out of $m$, not out of the full corpus.  This relative scoping is what makes iterative refinement informative:
\begin{itemize}\setlength{\itemsep}{1pt}
\item \emph{Token decomposition}: for each token ($\geq$4 chars) extracted from the \emph{regex string} (via \verb|[^\W_]{4,}|), the number of documents in the target set whose normalized text matches that token as a standalone, case-insensitive regex probe.  This reveals which token is the bottleneck (too strict) or which is vacuous (matches everything).
\item \emph{Keyword intersection}: the number of documents containing \emph{both} the top two keywords extracted from $\rho$ simultaneously, revealing how much the combined pattern over-constrains.
\item \emph{Selectivity gap}: when the keyword intersection is larger than the regex match count, the system flags: ``$k$ docs likely use a different format.  Read negatives.''  This prompts the agent to investigate format variants it is missing.
\item \emph{Anchor-token coverage} (\texttt{entity\_coverage} in logs): for the question's primary anchor token (see Appendix \ref{app:context}), how many documents containing that token fall inside vs.\ outside the current set.
\item \emph{Negative anchor-token check}: how many negative (excluded) documents still contain the top question-derived anchor token, alerting the agent to potential false negatives.
\item \emph{Scope check}: when all docs in a narrow subset match, the system checks how many docs \emph{outside} that subset also match the same regex, preventing premature narrowing.
\item \emph{Positive and negative snippets}: text excerpts from both sides of the partition, centered on the regex match (positive) or the top anchor token (negative), so the agent can see \emph{why} a document was included or excluded.
\end{itemize}

\paragraph{\texttt{count\_set} diagnostics.}
Beyond the count itself:
\begin{itemize}\setlength{\itemsep}{1pt}
\item \emph{Provenance warning}: if the set was never refined by a condition-specific regex, the system warns ``unrefined set---count may include false positives.''
\item \emph{Anchor-token coverage}: how many docs containing the top anchor token are outside the counted set.
\item \emph{Filename range}: the lexicographically first and last filenames, showing the scope of the set at a glance.
\end{itemize}

\paragraph{\texttt{extract\_field} / \texttt{extract\_list} diagnostics.}
The observation reports: how many documents were scanned, how many yielded a value, and a sample of extracted values, letting the agent immediately see whether its regex captures the intended structure or misses most documents.

\medskip
% \noindent These diagnostics are not post-hoc analysis tools, they are \emph{part of every observation}, computed automatically, and directly visible to the LLM in the conversation.  They close the gap between ``the agent tried something'' and ``the agent knows whether it worked,'' enabling self-correction without additional tool calls.

\begin{figure}[t]
\centering
\fbox{\parbox{0.95\columnwidth}{
\textbf{Example: multi-hop question over MuSiQue.}  ``What county is Erik Hort's birthplace a part of?'' 
% (2 hops, 11{,}656 docs)
\smallskip

\texttt{Step 1: apply\_filter("Erik", all)} $\to$ \texttt{has\_Erik\_pos} (30 docs).\smallskip

\texttt{Step 2: apply\_filter("Hort", has\_Erik\_pos)} $\to$ 2 docs, including \emph{Erik\_Hort.txt}.  Observation snippets reveal birthplace = Montebello, NY.\smallskip

\texttt{Step 3: apply\_filter("Montebello", all)} $\to$ 3 docs.  Observation includes \emph{Montebello\_New\_York.txt}: ``incorporated village in\ldots\ Rockland County.''

\medskip\noindent\textbf{Answer:} Rockland County.  (3 tool calls, correct.)
}}
\caption{Set-algebraic multi-hop reasoning.  Each hop narrows the set, and the agent uses the intermediate set as both evidence and anchor for the next hop.}
\label{fig:multihop}
\end{figure}

% \subsubsection{Formal Semantics of Persistence}
% \label{sssec:persistence}

% We now formalize the workspace dynamics.  At each step the agent selects a tool $\tau_t$ and arguments $\alpha_t$; the tool reads from and writes to the shared state:
% \begin{equation}
% \label{eq:step}
% (\mathcal{S}^{(t)}, \mathcal{V}^{(t)}, o_t) = \tau_t(\alpha_t; \mathcal{S}^{(t-1)}, \mathcal{V}^{(t-1)}, \mathcal{C})
% \end{equation}
% and the observation $o_t$ is appended to the conversation.  The agent's ``thought'' at turn $t$ conditions on all prior observations (possibly redacted) and the structural metadata of all named sets.  This creates a feedback loop: the agent sees what changed, and plans the next action accordingly.

% \paragraph{Why set persistence matters.}
% Consider the question ``In 2022-Q2, which company's earnings call had the most analyst participants?''  Without persistence, an agent would need to retrieve and compare analyst counts across all 14 calls in a single context window.  With persistence, the agent:
% (1) creates \texttt{q2\_2022\_calls} via filename search (14 docs),
% (2) reads a sample to learn the ``Call Participants'' section format,
% (3) extracts analyst counts into a value set, and
% (4) aggregates with $\mathrm{max}$.
% Each step builds on the previous, and the intermediate sets are available for the agent to inspect, refine, or reuse.

% \subsubsection{The Agent Loop}
% \label{sssec:loop}

%=======================================================================
\subsection{Limited-Ingestion ScalableRAG}
\label{ssec:limited}
%=======================================================================

Zero-Ingestion ScalableRAG deliberately pays no preprocessing cost, so when it creates a set out of a bigger set it uses regex on the entire set of documents of the bigger set.  This is the preferable operating point
when the corpus is small enough that a scan is cheap.  Limited-Ingestion ScalableRAG preserves
the same persistent state registry and downstream algebra, but restricts how
new subsets may be proposed while the parent set is large.% We remark that the set registry workspace is
% essential to this regime: indexed retrieval materializes a named candidate
% set only when its complete extension is available, and subsequent steps operate on that subset rather than restarting from
% the full corpus or serializing its contents into the LLM context.

% This changes which quantity dominates execution. A Zero-Ingestion filter scans
% its complete parent set, whereas Limited-Ingestion first uses the index to
% materialize a candidate relation and makes later work depend on the progressively
% smaller active set. Because each result remains a named operand, narrowing is
% cumulative: the agent pays large-corpus access to construct a set, then reuses
% that set for subsequent filters, intersections, extractions, and reductions.
% An index alone can return matching text or identifiers; supporting these later
% complete-set operations requires materializing those identifiers as a reusable
% relation, which is precisely the workspace abstraction.

\paragraph{Tiered filtering.}
A set is considered small only when it contains at most 1{,}000
documents and at most 10 MB of text.  On a larger set, the agent may issue
literal queries (including literal disjunctions) against a lexical index, but not an arbitrary regex. This is because with a lexical index literals and literal disjunctions cost time sublinear in corpus size, and are therefore much cheaper at big scales.
Only a complete result within the materialization limit forms another named set; if the result would remain large, the operation returns a bounded diagnostic
preview but performs no registry transition, and the agent must issue a more
selective query against the parent set. Once an accepted set crosses the threshold (<1000 documents and less than 10MB), the full
Zero-Ingestion regex interface becomes available again.  Thus ingestion
changes the cost of proposing and refining persistent sets, but not the
set-algebraic reasoning that follows. See Appendix~\ref{app:limited-details} for details.
% Large-set literals use the lexical
% index's token semantics; they are not treated as a recall-preserving surrogate
% for arbitrary substring regexes. 

% \paragraph{Cost boundary.}
% Let $M$ be the number of documents in an accepted indexed result. Once that
% result is materialized, downstream filtering, set algebra, extraction, and
% reduction operate on the named operands and scale with their touched
% cardinalities rather than repeatedly with corpus size $N$. This is the relevant
% large-$N$ regime: a selective set can be constructed in a million-document
% corpus and then aggregated under the same interface. If the desired set itself
% has $M=\Theta(N)$ previously unseen members, any exact method must inspect or
% precompute information about those members, incurring corresponding work;
% ScalableRAG pays that unavoidable cost only for question-relevant operands
% while avoiding corpus-wide preprocessing entirely.

\paragraph{Additional ingestion.}
Since this regime already accepts preprocessing, we also introduce a lightweight (constant number of LLM calls in corpus size) discovery and caching of regex patterns.  Pattern discovery uses deterministic scanners for recurring fields
and sections together with LLM-proposed templates.  For each retained pattern,
the ingestion pass indexes both matching document identifiers and bounded
captures. Consequently, \texttt{pattern\_search} uses an indexed identifier
lookup rather than rescanning document text, avoiding the full tiered filtering
path for cached patterns. We further leave open the option to use coarse vector embeddings (\verb|text-embedding-3-small| chunks in keeping with the limited-ingestion design), but in practice only apply it in corpora under 100K documents, as it requires non-trivial ingestion time for corpora over 500K, and does not seem to affect performance much. See Appendix \ref{app:limited-details} for details.

% Consequently there are two additional manners to create new sets: semantic search via embeddings (top $k$), and pattern lookups. (See Appendix \ref{app:limited-details} for details.) Embeddings and patterns are useful fallbacks, not
% the conceptual distinction between the two regimes.  We therefore evaluate
% Limited-Ingestion as a configured system and use one targeted embedding
% ablation to test whether its scaling behavior depends on semantic retrieval;
% we do not claim a full component-wise attribution.

%=======================================================================
\section{Experiments}
\label{sec:experiments}
%=======================================================================

\subsection{Experimental Setup}
The evaluation characterizes the complete ScalableRAG system across complementary workloads. 

The four \textbf{conventional datasets} are: MuSiQue \cite{trivedi2022musique}, with 11{,}656 docs and 1{,}000 questions; 2WikiMultiHopQA \cite{ho2020constructing}, with 6{,}118 docs and 1{,}000 questions; FinanceBench \cite{islam2023financebench}, with 84 financial filing documents and 150 questions; and ComplexTR \cite{tan2024towards}, with 872 documents and 200 temporal multi-hop questions.

The \textbf{aggregative datasets} are: Hotels-50 with 50 hotel documents and 138 questions and Hotels-350 with 350 hotel documents and 193 questions \cite{koshorek2025structured}. We remark that the Hotels datasets were introduced in the SRAG paper, to show the benefit of SRAG for aggregative questions. In \cite{koshorek2025structured} the results are reported on Hotels-350. We also use Transcripts (a new corpus-operations dataset constructed from public U.S.\ Government Publishing Office (GPO) records of the 117th Congress;\footnote{Source documents from \url{https://www.govinfo.gov/app/collection/chrg/117}.} see Appendix~\ref{app:chrg}), with 75 U.S.\ Congressional hearing transcripts and 100 questions, which combines local and corpus-operation questions in one corpus.

The \textbf{scale datasets} are constructed as follows: we begin with the 6{,}118 2WikiMultiHopQA documents, and append
distinct full articles from the fixed \texttt{20231101.en} Wikipedia snapshot. We discard empty articles and exact SHA-256 duplicates, shuffle with a fixed seed, and complement to get 100K, 500K, 1M, and 5M documents. All conditions use the same fixed 100-question subset and gold answers from 2WikiMultiHopQA.

\paragraph{Protocol.}
All systems answer identical fixed question sets, run end to end with the same GPT-4.1 generation model, and are scored by the same two LLM judges and the same EM/F1 normalization (Appendix~\ref{app:judge}). ScalableRAG uses one dataset-agnostic prompt, tool set, and threshold configuration on every corpus (Appendix~\ref{app:prompt}); no parameter or corpus-specific rule is fit to any labels. Baselines run from their authors' released code and prompts with default settings, with only the generation model unified to GPT-4.1; each ingestion-heavy baseline builds its schema or graph from the corpus it answers over, as in its own paper, so SRAG is evaluated on Hotels-350 exactly as its authors evaluate it. Our SRAG score on Hotels-350 (46.63) is in line with the 51.8 its authors report for the same automatic inferred-schema configuration under their own GPT-4o judge; their gold-schema variant requires a human-written schema and is not comparable to systems with no corpus-specific input. ScalableRAG decodes at temperature 0. Judges see only the question, the gold answer, and the extracted final answer, never the system identity. Appendix~\ref{app:inference} reports a 95\% confidence interval for every completed system--dataset cell.

\subsection{Results}
Table~\ref{tab:main_results} shows the breadth of the resulting system. Across all seven corpora, Zero-Ingestion reaches a macro GPT-4.1 judge score of 72.67 with no ingestion at all, ahead of A-RAG (61.51), HippoRAG2 (56.50), Vanilla RAG (43.97), SRAG (28.69), and GraphRAG (18.04); Limited-Ingestion is highest overall (73.70).

On Hotels-50, the purely aggregative workload, Zero-Ingestion scores 71.74 with no preprocessing or vector index. This is 40.22 points above A-RAG and 36.23 above relational SRAG, which performs offline schema extraction; every ingestion-heavy structured system, each designed for such aggregation, scores at most 35.51. On Hotels-350, the larger partition on which SRAG reports its own results, both variants again lead: Limited-Ingestion scores 62.44 and Zero-Ingestion 55.96, against 46.63 for SRAG and 22.02 for A-RAG. Zero-Ingestion also gains 3.50 points on the mixed aggregative/non-aggregative Transcripts benchmark.

On the four conventional multi-hop and document-grounded benchmarks, the same Zero-Ingestion system matches A-RAG in mean score (74.75 versus 74.64). The two kinds of gap differ in kind, not just in size. On Hotels-50, Zero-Ingestion scores 71.74 (95\% CI $[64.20,79.28]$), compared with 31.52 ($[23.70,39.34]$) for A-RAG and 35.51 ($[27.43,43.59]$) for SRAG. On Hotels-350, it scores 55.96 ($[48.89,63.03]$), compared with 22.02 ($[16.14,27.90]$) for A-RAG and 46.63 ($[39.53,53.73]$) for SRAG. Gemini yields the same qualitative pattern; Appendix~\ref{app:inference} reports the interval for every system--dataset cell. We accordingly read the aggregative gains as clear and the conventional-QA differences as practical parity.

% Across all seven corpora, Zero-Ingestion consequently reaches a macro GPT-4.1 judge score of 72.67, ahead of A-RAG (61.51), HippoRAG2 (56.50), Vanilla RAG (43.97), SRAG (28.69), and GraphRAG (18.04); its query-weighted score is likewise higher than A-RAG's (71.90 versus 67.43). Limited-Ingestion attains the strongest macro and query-weighted averages overall (73.70 and 72.86). The structured baselines behave according to their design envelopes: GraphRAG retrieves graph- and community-level summaries, which rarely contain the per-document fields these questions require, and SRAG commits to a single inferred schema per corpus, which is effective where questions match the extracted columns (its strongest result, 74.00, is on the highly regular Transcripts corpus) and brittle otherwise (3.33--14.75 on open-domain Wikipedia QA). This is the central end-to-end result: 

Table~\ref{tab:main_results} therefore already shows that ScalableRAG occupies an operating point absent from the evaluated alternatives, combining the ordinary-QA quality of agentic retrieval with stronger aggregation than ingestion-heavy structured systems while requiring zero or limited materialization.

Table \ref{tab:scale} compares Vanilla RAG (top 100)\footnote{Vanilla RAG uses \texttt{all-MiniLM-L6-v2} with 900-character windows and 50\% overlap.}, Limited-Ingestion ScalableRAG, and A-RAG on the scale datasets; the other baselines' ingestion costs are too high to complete in reasonable time at these scales. The result is a new quality--ingestion frontier: A-RAG scores higher at 100K and 500K (88.50 and 84.00 versus 80.50 and 79.00), but its cost grows with the corpus: per-question latency rises from 27.93 to 104.25 to 213.31 seconds across 100K, 500K, and 1M, ingestion from 6{,}312 to 30{,}270 to 57{,}500 seconds, while its score at 1M is 80.00. Limited-Ingestion holds latency flat, so at 1M it scores 84.50 (95\% CI $[77.49,91.51]$), compared with 80.00 ($[72.28,87.72]$) for A-RAG, at 13.53 versus 213.31 seconds per question and 269.58 versus 57{,}500 seconds of ingestion. A-RAG's 5M index did not finish within 48 hours. At 5M, among systems completing within the 48-hour budget, Limited-Ingestion is the only one scoring above 80, with under 20 minutes of ingestion; dense Vanilla RAG completes its run but scores only 16.3.

\begin{table*}[t]

\centering

\scriptsize
{
\setlength{\tabcolsep}{0.65pt}\renewcommand{\arraystretch}{0.92}%
\newsavebox{\mainresultsbox}
\savebox{\mainresultsbox}{%
\begin{tabular}{@{}lcccccccc@{}}
\toprule
System & MuSiQue & 2Wiki & Transcripts & FinanceBench & ComplexTR & Hotels-50 & Hotels-350 & Aggregate\\
\midrule
\begin{tabular}{@{}l@{}}\textbf{Limited-Ingestion}\\[-1pt]\textbf{ScalableRAG}\end{tabular} & \begin{tabular}{@{}c@{}}\textbf{58.91}/\textbf{55.90}\\[-1pt]{\tiny \textbf{33.70}/\textbf{49.93}}\end{tabular} & \begin{tabular}{@{}c@{}}86.76/\textbf{85.70}\\[-1pt]{\tiny 59.70/73.69}\end{tabular} & \begin{tabular}{@{}c@{}}\textbf{83.50}/\textbf{77.50}\\[-1pt]{\tiny \textbf{31.00}/\textbf{59.41}}\end{tabular} & \begin{tabular}{@{}c@{}}75.58/\textbf{74.70}\\[-1pt]{\tiny \textbf{20.67}/\textbf{38.10}}\end{tabular} & \begin{tabular}{@{}c@{}}75.57/76.30\\[-1pt]{\tiny 60.00/75.56}\end{tabular} & \begin{tabular}{@{}c@{}}\textbf{73.15}/\textbf{72.50}\\[-1pt]{\tiny \textbf{26.09}/\textbf{35.28}}\end{tabular} & \begin{tabular}{@{}c@{}}\textbf{62.44}/\textbf{62.69}\\[-1pt]{\tiny \textbf{14.51}/\textbf{22.77}}\end{tabular} & \begin{tabular}{@{}c@{}}\textbf{73.70}/\textbf{72.86}\\[-1pt]{\tiny \textbf{35.10}/\textbf{50.68}}\end{tabular}\\
\begin{tabular}{@{}l@{}}\textbf{Zero-Ingestion}\\[-1pt]\textbf{ScalableRAG}\end{tabular} & \begin{tabular}{@{}c@{}}\textbf{58.09}/54.10\\[-1pt]{\tiny 34.00/48.64}\end{tabular} & \begin{tabular}{@{}c@{}}85.74/84.60\\[-1pt]{\tiny 59.60/72.55}\end{tabular} & \begin{tabular}{@{}c@{}}\textbf{82.00}/77.00\\[-1pt]{\tiny 23.00/52.76}\end{tabular} & \begin{tabular}{@{}c@{}}76.31/73.80\\[-1pt]{\tiny \textbf{19.33}/\textbf{37.55}}\end{tabular} & \begin{tabular}{@{}c@{}}78.84/77.30\\[-1pt]{\tiny 60.00/77.37}\end{tabular} & \begin{tabular}{@{}c@{}}\textbf{71.74}/\textbf{72.50}\\[-1pt]{\tiny \textbf{23.19}/\textbf{32.60}}\end{tabular} & \begin{tabular}{@{}c@{}}\textbf{55.96}/\textbf{56.37}\\[-1pt]{\tiny 13.99/\textbf{19.50}}\end{tabular} & \begin{tabular}{@{}c@{}}\textbf{72.67}/\textbf{71.90}\\[-1pt]{\tiny \textbf{33.30}/\textbf{48.71}}\end{tabular}\\
\midrule
HippoRAG2 & \begin{tabular}{@{}c@{}}50.43/47.7\\[-1pt]{\tiny 27.40/42.48}\end{tabular} & \begin{tabular}{@{}c@{}}73.99/73.50\\[-1pt]{\tiny 58.50/65.69}\end{tabular} & \begin{tabular}{@{}c@{}}76.00/\textbf{77.5}\\[-1pt]{\tiny 23.00/\textbf{53.52}}\end{tabular} & \begin{tabular}{@{}c@{}}61.08/62.3\\[-1pt]{\tiny 13.33/29.10}\end{tabular} & \begin{tabular}{@{}c@{}}\textbf{81.34}/\textbf{79.10}\\[-1pt]{\tiny \textbf{68.50}/82.20}\end{tabular} & \begin{tabular}{@{}c@{}}32.97/33.0\\[-1pt]{\tiny 6.52/10.54}\end{tabular} & \begin{tabular}{@{}c@{}}19.69/19.69\\[-1pt]{\tiny 0.52/3.02}\end{tabular} & \begin{tabular}{@{}c@{}}56.50/59.62\\[-1pt]{\tiny 28.25/40.94}\end{tabular}\\
SRAG & \begin{tabular}{@{}c@{}}3.33/2.80\\[-1pt]{\tiny 1.40/2.41}\end{tabular} & \begin{tabular}{@{}c@{}}14.75/14.20\\[-1pt]{\tiny 11.40/13.00}\end{tabular} & \begin{tabular}{@{}c@{}}74.00/74.00\\[-1pt]{\tiny \textbf{26.00}/41.01}\end{tabular} & \begin{tabular}{@{}c@{}}18.25/16.60\\[-1pt]{\tiny 1.33/7.27}\end{tabular} & \begin{tabular}{@{}c@{}}8.33/8.50\\[-1pt]{\tiny 5.50/7.11}\end{tabular} & \begin{tabular}{@{}c@{}}35.51/35.90\\[-1pt]{\tiny 16.67/17.47}\end{tabular} & \begin{tabular}{@{}c@{}}46.63/46.63\\[-1pt]{\tiny \textbf{14.51}/15.45}\end{tabular} & \begin{tabular}{@{}c@{}}28.69/15.74\\[-1pt]{\tiny 10.97/14.82}\end{tabular}\\
\begin{tabular}{@{}l@{}}Vanilla RAG\\[-1pt](top-100)\end{tabular} & \begin{tabular}{@{}c@{}}38.49/35.50\\[-1pt]{\tiny 21.30/31.30}\end{tabular} & \begin{tabular}{@{}c@{}}43.80/42.90\\[-1pt]{\tiny 25.90/35.03}\end{tabular} & \begin{tabular}{@{}c@{}}71.00/65.50\\[-1pt]{\tiny 19.00/47.23}\end{tabular} & \begin{tabular}{@{}c@{}}32.09/30.80\\[-1pt]{\tiny 4.00/14.96}\end{tabular} & \begin{tabular}{@{}c@{}}77.59/77.70\\[-1pt]{\tiny 64.50/77.88}\end{tabular} & \begin{tabular}{@{}c@{}}30.80/31.20\\[-1pt]{\tiny 2.90/10.35}\end{tabular} & \begin{tabular}{@{}c@{}}13.99/13.47\\[-1pt]{\tiny 2.59/4.91}\end{tabular} & \begin{tabular}{@{}c@{}}43.97/41.95\\[-1pt]{\tiny 20.03/31.67}\end{tabular}\\
A-RAG & \begin{tabular}{@{}c@{}}57.00/53.60\\[-1pt]{\tiny 31.80/48.04}\end{tabular} & \begin{tabular}{@{}c@{}}\textbf{87.05}/85.60\\[-1pt]{\tiny \textbf{61.10}/\textbf{74.64}}\end{tabular} & \begin{tabular}{@{}c@{}}78.50/71.00\\[-1pt]{\tiny 18.00/45.97}\end{tabular} & \begin{tabular}{@{}c@{}}\textbf{77.41}/74.60\\[-1pt]{\tiny 18.00/34.84}\end{tabular} & \begin{tabular}{@{}c@{}}77.09/76.70\\[-1pt]{\tiny 60.00/75.88}\end{tabular} & \begin{tabular}{@{}c@{}}31.52/30.40\\[-1pt]{\tiny 7.25/12.12}\end{tabular} & \begin{tabular}{@{}c@{}}22.02/22.28\\[-1pt]{\tiny 3.63/6.03}\end{tabular} & \begin{tabular}{@{}c@{}}61.51/67.43\\[-1pt]{\tiny 28.54/42.50}\end{tabular}\\
GraphRAG & \begin{tabular}{@{}c@{}}15.47/6.70\\[-1pt]{\tiny 7.00/12.87}\end{tabular} & \begin{tabular}{@{}c@{}}35.55/19.00\\[-1pt]{\tiny 30.00/34.69}\end{tabular} & \begin{tabular}{@{}c@{}}35.50/14.00\\[-1pt]{\tiny 1.00/25.22}\end{tabular} & \begin{tabular}{@{}c@{}}23.54/5.90\\[-1pt]{\tiny 4.00/10.49}\end{tabular} & \begin{tabular}{@{}c@{}}13.75/12.70\\[-1pt]{\tiny 8.50/15.01}\end{tabular} & \begin{tabular}{@{}c@{}}1.45/0.70\\[-1pt]{\tiny 0.00/0.00}\end{tabular} & \begin{tabular}{@{}c@{}}1.04/2.59\\[-1pt]{\tiny 0.52/0.52}\end{tabular} & \begin{tabular}{@{}c@{}}18.04/22.03\\[-1pt]{\tiny 7.29/14.11}\end{tabular}\\
AutoSchemaKG & \begin{tabular}{@{}c@{}}---\\[-1pt]{\tiny ---}\end{tabular} & \begin{tabular}{@{}c@{}}---\\[-1pt]{\tiny ---}\end{tabular} & \begin{tabular}{@{}c@{}}---\\[-1pt]{\tiny ---}\end{tabular} & \begin{tabular}{@{}c@{}}---\\[-1pt]{\tiny ---}\end{tabular} & \begin{tabular}{@{}c@{}}80.40/79.60\\[-1pt]{\tiny 67.50/\textbf{82.41}}\end{tabular} & \begin{tabular}{@{}c@{}}24.64/25.40\\[-1pt]{\tiny 2.17/4.97}\end{tabular} & \begin{tabular}{@{}c@{}}22.80/22.54\\[-1pt]{\tiny 1.04/3.37}\end{tabular} & \begin{tabular}{@{}c@{}}---\\[-1pt]{\tiny ---}\end{tabular}\\
\bottomrule
\end{tabular}}
\ifdim\wd\mainresultsbox>\textwidth\resizebox{\textwidth}{!}{\usebox{\mainresultsbox}}\else\usebox{\mainresultsbox}\fi}
\caption{Main results (\%). Dataset cells report mean GPT-4.1 / Gemini~2.5 judge score (top) and EM/F1 (bottom). Hotels-50 and Hotels-350 are the two public partitions of the Hotels benchmark (50 and 350 documents); Hotels-350 is the corpus on which SRAG reports its own results. Aggregate cells report macro / query-weighted GPT-4.1 judge score (top) and macro EM/F1 (bottom) over all seven corpora. Judge scores permit fractional credit and are averaged over questions (Appendix~\ref{app:judge}); per-cell confidence intervals are in Appendix~\ref{app:inference}. Systems used GPT-4.1 on a GCP n1-standard-8 instance. Limited-Ingestion uses coarse embeddings on these corpora and none in Table~\ref{tab:scale}. Dashes denote unavailable results.
}
\label{tab:main_results}
\end{table*}

% \begin{table*}[t]
% \centering
% \scriptsize
% \setlength{\tabcolsep}{2.6pt}
% \begin{tabular}{rcrrrrrrrr}
% \toprule
% Docs & GB & \multicolumn{2}{c}{Zero} & \multicolumn{2}{c}{Limited} & \multicolumn{2}{c}{Vanilla RAG (top 100)} & \multicolumn{2}{c}{A-RAG} \\
% \cmidrule(lr){3-4}\cmidrule(lr){5-6}\cmidrule(lr){7-8}\cmidrule(lr){9-10}
%  & & seconds & accuracy & seconds & accuracy & seconds & accuracy & seconds & accuracy \\
% \midrule
% 100K & 0.22 & 100.58 & 80.5 & 15.52 & 84.0 & 5.72 & 28.5 & 27.93 & \textbf{88.5} \\
% 500K & 1.48 & 515.10 & 82.5 & 14.65 & \textbf{81.5} & 8.20 & 21.0 & -- & -- \\
% 1M & 3.66 & -- & -- & 12.46 & \textbf{82.5} & 6.64 & 24.0 & -- & -- \\
% 5M & 14.79 & -- & -- & 20.04 & \textbf{84.5} & -- & -- & -- & -- \\
% \bottomrule
% \end{tabular}
% \caption{End-to-end scale study on nested 2Wiki+Wikipedia corpora (100 fixed
% questions).  Each pair reports mean online seconds per question and
% LLM-judge accuracy using GPT4.1 (\%).  Latency includes retrieval, tool execution, and answer generation, but excludes preprocessing and judging.  Limited denotes
% its default configuration, including embeddings.  Dashes denote cells not
% completed by the fixed measurement cutoff; we do not report partial-run
% accuracy. Appendix~\ref{app:scale-details} reports preprocessing and coverage.}
% \label{tab:scale}
% \end{table*}
\begin{table*}[t]
\centering
\scriptsize
{
\setlength{\tabcolsep}{0.65pt}
\renewcommand{\arraystretch}{0.92}%

\def\scorehead{%
  \begin{tabular}{@{}c@{}}
  judge score\\[-1pt]
  {\tiny EM/F1}
  \end{tabular}}

\def\latencyhead{%
  \begin{tabular}{@{}c@{}}
  inference (s)\\[-2pt]
  {\tiny ingestion (s)}
  \end{tabular}}

\begin{tabular}{@{}lcrcrcrcr@{}}
\toprule
& \multicolumn{2}{c}{100K}
& \multicolumn{2}{c}{500K}
& \multicolumn{2}{c}{1M}
& \multicolumn{2}{c}{5M}\\
\cmidrule(lr){2-3}
\cmidrule(lr){4-5}
\cmidrule(lr){6-7}
\cmidrule(lr){8-9}
System
& \scorehead & \latencyhead
& \scorehead & \latencyhead
& \scorehead & \latencyhead
& \scorehead & \latencyhead\\
\midrule

\begin{tabular}{@{}l@{}}
\textbf{Limited-Ingestion}\\[-1pt]
\textbf{ScalableRAG}
\end{tabular}
&
\begin{tabular}{@{}c@{}}
80.50/81.25\\[-1pt]
{\tiny 51.00/64.29}
\end{tabular}
&
\begin{tabular}{@{}c@{}}
\textbf{16.45}\\[-1pt]
{\tiny \textbf{14.24}}
\end{tabular}
&
\begin{tabular}{@{}c@{}}
79.00/79.00\\[-1pt]
{\tiny 52.00/61.90}
\end{tabular}
&
\begin{tabular}{@{}c@{}}
\textbf{13.07}\\[-1pt]
{\tiny \textbf{109.08}}
\end{tabular}
&
\begin{tabular}{@{}c@{}}
\textbf{84.50}/\textbf{84.50}\\[-1pt]
{\tiny 49.00/\textbf{66.43}}
\end{tabular}
&
\begin{tabular}{@{}c@{}}
\textbf{13.53}\\[-1pt]
{\tiny \textbf{269.58}}
\end{tabular}
&
\begin{tabular}{@{}c@{}}
\textbf{81.50}/\textbf{78.50}\\[-1pt]
{\tiny \textbf{49.00}/\textbf{63.50}}
\end{tabular}
&
\begin{tabular}{@{}c@{}}
\textbf{13.64}\\[-1pt]
{\tiny \textbf{1{,}166.01}}
\end{tabular}
\\

\begin{tabular}{@{}l@{}}
\textbf{Zero-Ingestion}\\[-1pt]
\textbf{ScalableRAG}
\end{tabular}
&
\begin{tabular}{@{}c@{}}
80.50/83.00\\[-1pt]
{\tiny 52.00/65.31}
\end{tabular}
&
\begin{tabular}{@{}c@{}}
100.58\\[-1pt]
{\tiny 0.00}
\end{tabular}
&
\begin{tabular}{@{}c@{}}
82.50/\textbf{83.00}\\[-1pt]
{\tiny 54.00/66.21}
\end{tabular}
&
\begin{tabular}{@{}c@{}}
515.12\\[-1pt]
{\tiny 0.00}
\end{tabular}
&
\begin{tabular}{@{}c@{}}
82.50/82.50\\[-1pt]
{\tiny 50.00/64.92}
\end{tabular}
&
\begin{tabular}{@{}c@{}}
1{,}076.25\\[-1pt]
{\tiny 0.00}
\end{tabular}
&
\begin{tabular}{@{}c@{}}
---/---$^\dagger$\\[-1pt]
{\tiny ---/---}
\end{tabular}
&
\begin{tabular}{@{}c@{}}
4{,}746.79\\[-1pt]
{\tiny 0.00}
\end{tabular}
\\

\midrule

\begin{tabular}{@{}l@{}}
Vanilla RAG\\[-1pt]
(top-100)
\end{tabular}
&
\begin{tabular}{@{}c@{}}
28.50/27.00\\[-1pt]
{\tiny 17.00/22.15}
\end{tabular}
&
\begin{tabular}{@{}c@{}}
5.72\\[-1pt]
{\tiny 656.00}
\end{tabular}
&
\begin{tabular}{@{}c@{}}
21.00/20.00\\[-1pt]
{\tiny 10.00/13.92}
\end{tabular}
&
\begin{tabular}{@{}c@{}}
8.20\\[-1pt]
{\tiny 4{,}399.00}
\end{tabular}
&
\begin{tabular}{@{}c@{}}
24.00/22.50\\[-1pt]
{\tiny 13.00/17.52}
\end{tabular}
&
\begin{tabular}{@{}c@{}}
6.64\\[-1pt]
{\tiny 10{,}902.00}
\end{tabular}
&
\begin{tabular}{@{}c@{}}
16.30/16.00\\[-1pt]
{\tiny 9.00/11.37}
\end{tabular}
&
\begin{tabular}{@{}c@{}}
10.58\\[-1pt]
{\tiny 44{,}326.00}
\end{tabular}
\\

A-RAG
&
\begin{tabular}{@{}c@{}}
\textbf{88.50}/\textbf{85.50}\\[-1pt]
{\tiny \textbf{58.00}/\textbf{70.99}}
\end{tabular}
&
\begin{tabular}{@{}c@{}}
27.93\\[-1pt]
{\tiny 6{,}312.00}
\end{tabular}
&
\begin{tabular}{@{}c@{}}
\textbf{84.00}/82.00\\[-1pt]
{\tiny \textbf{55.00}/\textbf{66.50}}
\end{tabular}
&
\begin{tabular}{@{}c@{}}
104.25\\[-1pt]
{\tiny 30{,}270.00}
\end{tabular}
&
\begin{tabular}{@{}c@{}}
80.00/79.25\\[-1pt]
{\tiny \textbf{54.00}/65.39}
\end{tabular}
&
\begin{tabular}{@{}c@{}}
213.31\\[-1pt]
{\tiny 57{,}500.00}
\end{tabular}
&
\begin{tabular}{@{}c@{}}
---/---$^\dagger$\\[-1pt]
{\tiny ---/---}
\end{tabular}
&
\begin{tabular}{@{}c@{}}
---\\[-1pt]
{\tiny $>172{,}800.00$}
\end{tabular}
\\

\bottomrule
\end{tabular}
}

\caption{Scale-study results on nested 2Wiki+Wikipedia corpora. Column groups
give corpus size. Score cells report mean GPT-4.1/Gemini~2.5 judge score (\%)
on the top line and EM/F1 (\%) on the bottom line. Latency cells report mean
end-to-end inference time (model, retrieval, and tool execution; judging
excluded) per question on the top line and durable document-store population
plus FTS5 index-construction latency on the bottom line, both in seconds. 
Bold denotes the highest score for each judge and lowest inference latency
for a non-Vanilla RAG system, and the lowest latency for a non-zero-ingestion
system. All systems ran on CPU-only instances. Dashes denote results unavailable at the measurement cutoff.
$^\dagger$ These runs did not finish in 48 hours.
Confidence intervals for all completed scale cells are reported in Appendix~\ref{app:inference}.
}
\label{tab:scale}
\end{table*}
\paragraph{Ablation: the workspace carries the aggregative gain.}
To isolate the mechanism, we ran Zero-Ingestion on Hotels-50 with the persistent-set workspace removed while retaining the regex retrieval tools, so intermediate subsets can no longer be stored, named, or reused across steps. The judge score falls from 71.74 to 38.4, into the range of the transient-state baselines (A-RAG 31.52, SRAG 35.51). The retrieval primitives alone therefore do not explain the gains on aggregative QA; retaining and composing operations on top of the set registry does. Nor is the gain purchased with longer trajectories: Zero-Ingestion and A-RAG operate at comparable agent depth, with 3.42--8.07 versus 3.26--7.63 mean steps per question across six corpora (all but Hotels-350). This matched comparison, together with the per-workload pattern above (largest gains precisely where corpus operations are required, parity elsewhere), ties the empirical results to the stated abstraction.

\section{Conclusion}
ScalableRAG closes the capability--ingestion gap rather than choosing one side of it. Zero-Ingestion starts from raw documents, matches strong agentic RAG on conventional QA, and beats every evaluated baseline (agentic, graph, and schema) on aggregative QA (for example over 36 points on Hotels-50); removing its workspace returns performance to the transient-state range, tying the gain to the mechanism. Limited-Ingestion carries the same set interface through a 5M-document search space, maintaining 13 to 16 second inference with under 20 minutes of ingestion, matches A-RAG's quality at 1M documents with 16$\times$ lower latency and $0.5\%$ of its ingestion time, and is the only system to finish at 5M within the 48-hour budget with judge score above 80. In sum, one architecture spans regimes previously served by different systems: high-quality local retrieval, exact aggregation over complete question-specific subsets, and million-document access with zero or limited ingestion. No evaluated alternative occupies this operating point.
\section{Limitations}
A core design choice is that every workspace relation identifies a subset of documents. This supports direct aggregation when the grouping key corresponds to documents, as in the workloads evaluated here. It is less natural when one document contains many independently aggregated entities or when answering requires many-to-many joins across repeated entities; knowledge-graph or multi-table approaches may be preferable in those regimes. Set operations are deterministic relative to the predicates supplied by the agent, but diagnostics cannot guarantee that an LLM-generated predicate captures every semantic match. Our cost claims concern ingestion and large-corpus access latency (Table~\ref{tab:scale}); on the six capability corpora we compare agent depth but make no claim of lower per-question inference cost than agentic baselines.

\section{Ethical Considerations}
All existing corpora used in our experiments are publicly available under permissive licenses.  The Transcripts dataset is constructed from U.S.\ Government Publishing Office releases (\texttt{govinfo.gov}), which place the underlying hearing transcripts in the public domain.  Our experiments do not involve human subjects, and we are not publishing ScalableRAG as a pre-trained model, and it therefore makes no use of personally identifying information.  As with all LLM-based systems, the outputs depend on the underlying language model; downstream users should treat answers as derived evidence rather than authoritative claims. Any artifacts generated by AI (Cursor/ChatGPT/Claude) as part of this paper's code generation, ideation, or paper editing had been under manual supervision and verification by the authors. 

\bibliography{references2}

\appendix

\section{Per-Cell Confidence Intervals}
\label{app:inference}

Tables~\ref{tab:main-ci-a}--\ref{tab:scale-intervals} report a separate 95\%
confidence interval for each completed system--dataset cell, without combining
systems or datasets.  Judge score, F1, and per-question latency use
Student-$t$ intervals for the mean; EM uses a Wilson binomial interval.
Ingestion time is one durable document-store-plus-FTS5 run per cell, so
Table~\ref{tab:scale} reports it as a point measurement without a confidence
interval.

\begin{table*}[t]
\centering
\scriptsize
\setlength{\tabcolsep}{3.5pt}
\resizebox{\textwidth}{!}{%
\begin{tabular}{@{}llcccc@{}}
\toprule
System & Dataset & GPT-4.1 [95\% CI] & Gemini~2.5 [95\% CI] & EM [95\% CI] & F1 [95\% CI] \\
\midrule
Limited & MuSiQue & 58.91 [55.93, 61.89] & 55.90 [52.91, 58.89] & 33.70 [30.84, 36.69] & 49.93 [47.24, 52.62] \\
Limited & 2Wiki & 86.76 [84.73, 88.79] & 85.70 [83.59, 87.81] & 59.70 [56.63, 62.70] & 73.69 [71.37, 76.01] \\
Limited & Transcripts & 83.50 [76.30, 90.70] & 77.50 [69.99, 85.01] & 31.00 [22.78, 40.63] & 59.41 [52.37, 66.45] \\
Limited & FinanceBench & 75.58 [69.06, 82.10] & 74.70 [68.18, 81.22] & 20.67 [14.96, 27.84] & 38.10 [32.06, 44.14] \\
Limited & ComplexTR & 75.57 [70.38, 80.76] & 76.30 [71.20, 81.40] & 60.00 [53.08, 66.54] & 75.56 [70.64, 80.48] \\
Limited & Hotels-50 & 73.15 [65.67, 80.63] & 72.50 [65.02, 79.98] & 26.09 [19.48, 33.99] & 35.28 [27.93, 42.63] \\
Limited & Hotels-350 & 62.44 [55.60, 69.28] & 62.69 [55.84, 69.54] & 14.51 [10.23, 20.17] & 22.77 [17.37, 28.17] \\
Zero & MuSiQue & 58.09 [55.09, 61.09] & 54.10 [51.09, 57.11] & 34.00 [31.13, 36.99] & 48.64 [45.92, 51.36] \\
Zero & 2Wiki & 85.74 [83.64, 87.84] & 84.60 [82.43, 86.77] & 59.60 [56.53, 62.60] & 72.55 [70.18, 74.92] \\
Zero & Transcripts & 82.00 [74.47, 89.53] & 77.00 [69.35, 84.65] & 23.00 [15.84, 32.15] & 52.76 [45.42, 60.10] \\
Zero & FinanceBench & 76.31 [69.81, 82.81] & 73.80 [67.19, 80.41] & 19.33 [13.81, 26.38] & 37.55 [31.60, 43.50] \\
Zero & ComplexTR & 78.84 [74.08, 83.60] & 77.30 [72.44, 82.16] & 60.00 [53.08, 66.54] & 77.37 [72.81, 81.93] \\
Zero & Hotels-50 & 71.74 [64.20, 79.28] & 72.50 [65.02, 79.98] & 23.19 [16.93, 30.90] & 32.60 [25.44, 39.76] \\
Zero & Hotels-350 & 55.96 [48.89, 63.03] & 56.37 [49.32, 63.42] & 13.99 [9.80, 19.59] & 19.50 [14.25, 24.75] \\
HippoRAG2 & MuSiQue & 50.43 [47.44, 53.42] & 47.70 [44.70, 50.70] & 27.40 [24.73, 30.25] & 42.48 [39.82, 45.14] \\
HippoRAG2 & 2Wiki & 73.99 [71.33, 76.65] & 73.50 [70.81, 76.19] & 58.50 [55.42, 61.52] & 65.69 [62.93, 68.45] \\
HippoRAG2 & Transcripts & 76.00 [67.60, 84.40] & 77.50 [69.48, 85.52] & 23.00 [15.84, 32.15] & 53.52 [45.49, 61.55] \\
HippoRAG2 & FinanceBench & 61.08 [53.51, 68.65] & 62.30 [54.77, 69.83] & 13.33 [8.80, 19.69] & 29.10 [23.49, 34.71] \\
HippoRAG2 & ComplexTR & 81.34 [76.79, 85.89] & 79.10 [74.33, 83.87] & 68.50 [61.77, 74.54] & 82.20 [77.94, 86.46] \\
HippoRAG2 & Hotels-50 & 32.97 [25.06, 40.88] & 33.00 [25.09, 40.91] & 6.52 [3.47, 11.93] & 10.54 [5.82, 15.26] \\
HippoRAG2 & Hotels-350 & 19.69 [14.03, 25.35] & 19.69 [14.03, 25.35] & 0.52 [0.09, 2.88] & 3.02 [1.13, 4.91] \\
SRAG & MuSiQue & 3.33 [2.27, 4.39] & 2.80 [1.79, 3.81] & 1.40 [0.84, 2.34] & 2.41 [1.58, 3.24] \\
SRAG & 2Wiki & 14.75 [12.56, 16.94] & 14.20 [12.03, 16.37] & 11.40 [9.58, 13.52] & 13.00 [10.97, 15.03] \\
SRAG & Transcripts & 74.00 [65.25, 82.75] & 74.00 [65.25, 82.75] & 26.00 [18.40, 35.37] & 41.01 [32.50, 49.52] \\
SRAG & FinanceBench & 18.25 [12.24, 24.26] & 16.60 [10.88, 22.32] & 1.33 [0.36, 4.73] & 7.27 [4.36, 10.18] \\
SRAG & ComplexTR & 8.33 [4.56, 12.10] & 8.50 [4.66, 12.34] & 5.50 [3.10, 9.58] & 7.11 [3.76, 10.46] \\
SRAG & Hotels-50 & 35.51 [27.43, 43.59] & 35.90 [27.83, 43.97] & 16.67 [11.37, 23.77] & 17.47 [11.11, 23.83] \\
SRAG & Hotels-350 & 46.63 [39.53, 53.73] & 46.63 [39.53, 53.73] & 14.51 [10.23, 20.17] & 15.45 [10.42, 20.48] \\
\bottomrule
\end{tabular}
}
\caption{Per-cell 95\% confidence intervals for the first four systems in Table~\ref{tab:main_results}. Each interval is computed independently from that system's question-level results on that dataset.}
\label{tab:main-ci-a}
\end{table*}

\begin{table*}[t]
\centering
\scriptsize
\setlength{\tabcolsep}{3.5pt}
\resizebox{\textwidth}{!}{%
\begin{tabular}{@{}llcccc@{}}
\toprule
System & Dataset & GPT-4.1 [95\% CI] & Gemini~2.5 [95\% CI] & EM [95\% CI] & F1 [95\% CI] \\
\midrule
Vanilla & MuSiQue & 38.49 [35.54, 41.44] & 35.50 [32.60, 38.40] & 21.30 [18.87, 23.94] & 31.30 [28.71, 33.89] \\
Vanilla & 2Wiki & 43.80 [40.75, 46.85] & 42.90 [39.86, 45.94] & 25.90 [23.28, 28.70] & 35.03 [32.30, 37.76] \\
Vanilla & Transcripts & 71.00 [62.06, 79.94] & 65.50 [56.73, 74.27] & 19.00 [12.51, 27.78] & 47.23 [39.46, 55.00] \\
Vanilla & FinanceBench & 32.09 [24.85, 39.33] & 30.80 [23.75, 37.85] & 4.00 [1.85, 8.45] & 14.96 [10.86, 19.06] \\
Vanilla & ComplexTR & 77.59 [72.71, 82.47] & 77.70 [72.88, 82.52] & 64.50 [57.65, 70.80] & 77.88 [73.19, 82.57] \\
Vanilla & Hotels-50 & 30.80 [23.03, 38.57] & 31.20 [23.38, 39.02] & 2.90 [1.13, 7.22] & 10.35 [6.02, 14.68] \\
Vanilla & Hotels-350 & 13.99 [9.05, 18.93] & 13.47 [8.61, 18.33] & 2.59 [1.11, 5.92] & 4.91 [2.17, 7.65] \\
A-RAG & MuSiQue & 57.00 [54.03, 59.97] & 53.60 [50.59, 56.61] & 31.80 [28.99, 34.75] & 48.04 [45.40, 50.68] \\
A-RAG & 2Wiki & 87.05 [85.05, 89.05] & 85.60 [83.48, 87.72] & 61.10 [58.04, 64.07] & 74.64 [72.33, 76.95] \\
A-RAG & Transcripts & 78.50 [70.37, 86.63] & 71.00 [62.64, 79.36] & 18.00 [11.70, 26.67] & 45.97 [38.36, 53.58] \\
A-RAG & FinanceBench & 77.41 [71.09, 83.73] & 74.60 [68.15, 81.05] & 18.00 [12.68, 24.92] & 34.84 [29.14, 40.54] \\
A-RAG & ComplexTR & 77.09 [72.14, 82.04] & 76.70 [71.84, 81.56] & 60.00 [53.08, 66.54] & 75.88 [71.15, 80.61] \\
A-RAG & Hotels-50 & 31.52 [23.70, 39.34] & 30.40 [22.63, 38.17] & 7.25 [3.99, 12.83] & 12.12 [7.16, 17.08] \\
A-RAG & Hotels-350 & 22.02 [16.14, 27.90] & 22.28 [16.40, 28.16] & 3.63 [1.77, 7.30] & 6.03 [2.97, 9.09] \\
GraphRAG & MuSiQue & 15.47 [13.35, 17.59] & 6.70 [5.20, 8.20] & 7.00 [5.58, 8.75] & 12.87 [11.05, 14.69] \\
GraphRAG & 2Wiki & 35.55 [32.60, 38.50] & 19.00 [16.58, 21.42] & 30.00 [27.24, 32.91] & 34.69 [31.87, 37.51] \\
GraphRAG & Transcripts & 35.50 [26.01, 44.99] & 14.00 [7.23, 20.77] & 1.00 [0.18, 5.45] & 25.22 [19.12, 31.32] \\
GraphRAG & FinanceBench & 23.54 [16.92, 30.16] & 5.90 [2.32, 9.48] & 4.00 [1.85, 8.45] & 10.49 [6.67, 14.31] \\
GraphRAG & ComplexTR & 13.75 [9.17, 18.33] & 12.70 [8.35, 17.05] & 8.50 [5.37, 13.19] & 15.01 [10.59, 19.43] \\
GraphRAG & Hotels-50 & 1.45 [-0.57, 3.47] & 0.70 [-0.73, 2.13] & 0.00 [0.00, 2.71] & 0.00 [0.00, 0.00] \\
GraphRAG & Hotels-350 & 1.04 [-0.40, 2.48] & 2.59 [0.33, 4.85] & 0.52 [0.09, 2.88] & 0.52 [-0.50, 1.54] \\
AutoSchemaKG & ComplexTR & 80.40 [75.74, 85.06] & 79.60 [74.88, 84.32] & 67.50 [60.73, 73.61] & 82.41 [78.20, 86.62] \\
AutoSchemaKG & Hotels-50 & 24.64 [17.36, 31.92] & 25.40 [18.05, 32.75] & 2.17 [0.74, 6.19] & 4.97 [1.75, 8.19] \\
AutoSchemaKG & Hotels-350 & 22.80 [16.83, 28.77] & 22.54 [16.61, 28.47] & 1.04 [0.29, 3.70] & 3.37 [1.24, 5.50] \\
\bottomrule
\end{tabular}
}
\caption{Per-cell 95\% confidence intervals for the remaining systems in Table~\ref{tab:main_results}. Each interval is computed independently from that system's question-level results on that dataset.}
\label{tab:main-ci-b}
\end{table*}

\begin{table*}[t]
\centering
\scriptsize
\setlength{\tabcolsep}{3.5pt}
\resizebox{\textwidth}{!}{%
\begin{tabular}{@{}llccccc@{}}
\toprule
System & Scale & GPT-4.1 [95\% CI] & Gemini~2.5 [95\% CI] & EM [95\% CI] & F1 [95\% CI] & Inference s [95\% CI] \\
\midrule
Limited-Ingestion & 100K & 80.50 [72.79, 88.21] & 81.25 [73.71, 88.79] & 51.00 [41.35, 60.58] & 64.29 [56.02, 72.56] & 16.45 [14.19, 18.71] \\
Limited-Ingestion & 500K & 79.00 [71.00, 87.00] & 79.00 [71.13, 86.87] & 52.00 [42.32, 61.54] & 61.90 [53.24, 70.56] & 13.07 [11.62, 14.52] \\
Limited-Ingestion & 1M & 84.50 [77.49, 91.51] & 84.50 [77.49, 91.51] & 49.00 [39.42, 58.65] & 66.43 [58.59, 74.26] & 13.53 [11.82, 15.24] \\
Limited-Ingestion & 5M & 81.50 [73.95, 89.05] & 78.50 [70.49, 86.51] & 49.00 [39.42, 58.65] & 63.50 [55.40, 71.59] & 13.64 [11.93, 15.35] \\
Zero-Ingestion & 100K & 80.50 [72.79, 88.21] & 83.00 [75.78, 90.22] & 52.00 [42.32, 61.54] & 65.31 [57.08, 73.54] & 100.58 [88.45, 112.71] \\
Zero-Ingestion & 500K & 82.50 [75.12, 89.88] & 83.00 [75.78, 90.22] & 54.00 [44.26, 63.44] & 66.21 [58.00, 74.41] & 515.12 [443.41, 586.83] \\
Zero-Ingestion & 1M & 82.50 [75.12, 89.88] & 82.50 [75.26, 89.74] & 50.00 [40.38, 59.62] & 64.92 [56.82, 73.02] & 1{,}076.25 [936.42, 1{,}216.08] \\
Vanilla RAG & 100K & 28.50 [19.55, 37.45] & 27.00 [18.37, 35.63] & 17.00 [10.89, 25.55] & 22.15 [14.40, 29.91] & 5.72 [5.55, 5.89] \\
Vanilla RAG & 500K & 21.00 [13.00, 29.00] & 20.00 [12.28, 27.72] & 10.00 [5.52, 17.44] & 13.92 [7.56, 20.28] & 8.20 [7.55, 8.85] \\
Vanilla RAG & 1M & 24.00 [15.48, 32.52] & 22.50 [14.23, 30.77] & 13.00 [7.76, 20.98] & 17.52 [10.44, 24.59] & 6.64 [6.48, 6.80] \\
Vanilla RAG & 5M & 16.30 [9.05, 23.55] & 16.00 [8.83, 23.17] & 9.00 [4.81, 16.23] & 11.37 [5.31, 17.43] & 10.58 [10.39, 10.77] \\
A-RAG & 100K & 88.50 [82.54, 94.46] & 85.50 [78.84, 92.16] & 58.00 [48.21, 67.20] & 70.99 [63.11, 78.88] & 27.93 [23.39, 32.47] \\
A-RAG & 500K & 84.00 [76.97, 91.03] & 82.00 [74.60, 89.40] & 55.00 [45.24, 64.39] & 66.50 [58.14, 74.87] & 104.25 [93.10, 115.40] \\
A-RAG & 1M & 80.00 [72.28, 87.72] & 79.25 [71.40, 87.10] & 54.00 [44.26, 63.44] & 65.39 [56.98, 73.80] & 213.31 [190.61, 236.01] \\
\bottomrule
\end{tabular}
}
\caption{Question-sampling uncertainty for every completed quality cell in the scale study. Values are the means from Table~\ref{tab:scale} followed by 95\% confidence intervals. The latency interval width is estimated from per-question elapsed times and centered on the corresponding end-to-end mean reported in Table~\ref{tab:scale}. Ingestion is intentionally omitted because each cell contains a single durable document-store-plus-FTS5 measurement.}
\label{tab:scale-intervals}
\end{table*}

\section{Limited-Ingestion Implementation}
\label{app:limited-details}
This appendix specifies the large-set execution policy summarized in Section~\ref{ssec:limited}.

All scale-study corpora are stored in SQLite and indexed using FTS5 with the \texttt{unicode61} tokenizer, supporting efficient lookup of individual terms and disjunctions of terms.

\subsection{Large-Set and Small-Set Semantics}

Let $S$ be the active document set.  The default small-set predicate is
$|S|\leq 1{,}000$ and \textsc{Bytes}$(S)\leq 10$ MB.  If the predicate holds,
\texttt{apply\_filter} evaluates the complete user regex over every document in
$S$. Otherwise, the operation accepts only a literal or a disjunction of
literals. Queries whose complete result would remain non-small are refused as
set-producing transitions. The agent is asked to issue a more selective query
against the same exact parent, narrow by filename, intersect existing exact
sets, use a cached pattern, or use semantic retrieval. Complex regexes are not
exposed for large sets.

On a large set, a literal is defined under the \texttt{unicode61} index's token
semantics, rather than substring-regex semantics. If the complete token-match
result is small, the literal regex is additionally evaluated on those candidate
texts, removing token-query false positives. This produces exact regex
membership relative to the token-defined candidate set. Full substring-regex
semantics apply after the workspace has narrowed to a small set. When a token
query would exceed the materialization limit, the tool may return a bounded
preview labeled \texttt{exact=false}. This preview is an observation, not a
document set: it is not inserted into $\mathcal{S}^{(t)}$ and cannot be used as
a parent, intersected, subtracted, extracted from, or counted. Refinement is a
new query over the unchanged exact parent, so identifiers omitted from a preview
never disappear from the lineage of a subsequently accepted set. Thus every
document set in the registry is the complete extension of its declared
operation over a complete parent. Negative sets are
represented lazily as a base set
minus the positive identifiers, avoiding materialization of millions of IDs.
The same index serves question-landscape counts, filename counts, filename
search, and deterministic Branch-and-Select previews.  Filename candidates are
retrieved through the indexed title field and then checked against the actual
filename, so indexing does not change match semantics.

\subsection{Pattern and Embedding Indexes}

Pattern discovery has three sources.  A deterministic scanner identifies
recurring line-initial \texttt{Label: Value} fields; a second deterministic
scanner identifies recurring Markdown sections; and an LLM reads a stable
hash-based sample of 20 documents and proposes parameterized structural
templates and semantic extraction hints.   LLM templates are sandboxed, evaluated on sampled matches and
non-matches, and refined for at most eight rounds.  Malformed or empty discovery
responses are retained in a diagnostic record and retried once rather than
silently becoming an empty pattern library.

For each retained regex template, ingestion stores its matching document IDs
and bounded named-group captures.  At inference, \texttt{pattern\_search}
intersects those IDs with the target set and evaluates parameter constraints
against the cached captures.  It therefore works on a full-corpus target
without rereading documents or re-executing regex.   

Coarse document embeddings are optional and add \texttt{cosine\_search}, which
requires a parameter $k$ and produces a ranked candidate relation of the top
$k$ document hits. Its provenance records the ranking operator and $k$; its
cardinality is not an exhaustive count of documents satisfying the natural-
language query. The agent prompt therefore uses this relation to inspect
examples and derive an exhaustive lexical or structural predicate before any
corpus-level count or aggregation.

\section{System Prompt}
\label{app:prompt}

The following is the complete system prompt provided to the ScalableRAG agent, copied verbatim from the source code.  Sections between \texttt{<<IF:X>>} and \texttt{<<ENDIF:X>>} are conditionally included depending on which modules are enabled (embeddings, pattern\_search, aggregation tools, and tiered filtering).  Template variables (\texttt{\{n\_docs\}}, \texttt{\{total\_chars\}}, etc.)\ are filled at runtime.

\begin{lstlisting}
You are a document retrieval agent.  You have a corpus of {n_docs} documents ({total_chars:,} characters).  Answer the question using ONLY evidence from the documents.  Never guess.

Context budget: ~{context_budget} tokens.  Previous observations may be redacted to save space -- your thought is preserved, so always record key findings in your thought.

=======================================================================
STEP 0 -- DECOMPOSE THE QUESTION
=======================================================================

Before doing ANYTHING, analyze the question for complexity:

* Does the question reference entities INDIRECTLY ("the X who...", "the Y where...")?  Each indirect reference is a HOP -- a fact you must resolve before you can answer.
* Count the hops.  If there are N hops, decompose the question into N+1 numbered sub-questions in your FIRST thought.
* After resolving each sub-question, state: "Sub-Q K resolved: [result].  Remaining: Sub-Q [K+1, ...]."
* Budget at least 2 tool calls per hop.

*** Your final answer must address the OUTERMOST question, not an intermediate sub-question.  If you resolved an intermediate fact but the question asks about a PROPERTY of that fact, you are NOT done -- keep searching. ***

=======================================================================
APPROACH -- for each sub-question (or the whole question if single-hop)
=======================================================================

1. START BROAD -- one keyword at a time.  Your first apply_filter MUST use a SINGLE entity keyword, never a compound pattern with multiple terms.  For structural constraints (dates, categories, document types), prefer search_filenames -- filenames encode metadata reliably while text mentions can be incidental.  Combine sets with set_operation(intersect).  False positives are fine -- false negatives are hard to recover.

2. EXPLORE from BOTH ENDS.  read_docs from the beginning AND end of your set (use offset=-3).  Documents are sorted by filename, so beginning/end often show different document types.  Note ALL format variants you observe -- a general category may appear in several distinct formats.  Build your refinement to cover ALL of them.

3. REFINE (mandatory for counting/aggregation).  "Contains keyword" != "Satisfies the condition."  Write a more specific regex targeting the actual format you observed -- anchor to structural markers (section headings, field labels), not bare keywords.  Apply WITHIN your working set.  If the negative set has docs in a different format, build a separate filter and UNION.

<<IF:tiered_filters>>
SCALING CONTRACT. On a large working set, apply_filter is index-backed and accepts only a literal keyword or a literal alternation such as term_a|term_b. Large-set literals use index token semantics; they are not arbitrary substring regexes. Complex regexes (.*, anchors, groups, character classes) are available only after the working set is small. If a complex filter is refused, first create a complete smaller set with a selective literal, search_filenames, or an intersection of exact sets, then retry the complex regex. An observation marked exact=false is only a bounded preview: it is not a workspace set and MUST NOT be filtered, intersected, extracted from, or counted. Reissue a more selective query against the last exact parent. Only exact results receive reusable set identity.
<<ENDIF:tiered_filters>>

<<IF:agg_tools>>
4. NUMERICAL AGGREGATION.  For questions asking for average, sum, min, max, or total of a value: first narrow to the exact relevant subset of docs (not "all" unless the question is about all docs). Then read a few docs to learn the exact format of the value.  Then use extract_field with a regex capture group to scan EVERY doc in that subset -- never compute extrema from a sample or from search results alone.  Check the "extracted" count matches your expectation.  Finally, aggregate with the right operation.  For min/max questions that ask about a PROPERTY of the extremal doc, use create_set to get the doc, then read_docs.  For dates, use as_type="date" so min/max give earliest/latest.

4b. LIST EXTRACTION.  When a question asks "how many X per doc", "who are the X", "list all X", or requires collecting repeating items (names, line items, participants, references) from each document, use extract_list -- NOT extract_field.  extract_field returns ONE value per doc; extract_list returns ALL matches as a list per doc.  Read a few docs first to learn the exact line format, write a regex with a capture group that matches each item, then run extract_list on the target set.  Follow with aggregate using a list operation: unique_values (deduplicated list), count_unique (number of distinct items), value_counts (frequency table), or group_by_value (which docs share each value).  Use these for cross-document questions like "which X appear in both A and B" or "how many unique X across all docs".
<<ENDIF:agg_tools>>

5. VERIFY & ANSWER.  For counts: read 2-3 docs from your NEGATIVE (excluded) set to verify your refinement regex did not miss valid variants.  If you find a different format, build a second filter and UNION.  Only then count_set on the refined set (not the broad set -- a broad count is an upper bound, never the answer).  For facts: quote exact wording from the source.  Include ALL relevant structured fields.  Use the document's own words -- do not paraphrase.

=======================================================================
TOOLS
=======================================================================

apply_filter  -- Apply regex to a set.  Creates _pos and _neg sets.  Text is section-normalized: lines within a paragraph are joined (use .* to span within a section, \n for section boundaries).  The result includes a keyword_decomposition showing how many docs match each individual keyword -- use this to gauge selectivity and find a better strategy when your combined pattern is too strict.
  {{"filter_type": "regex", "pattern": "term", "target_set": "all", "output_prefix": "has_term"}}
  {{"filter_type": "regex", "pattern": "precise", "target_set": "broad_pos", "output_prefix": "refined"}}
<<IF:embeddings>>
  {{"filter_type": "classifier", "filter_id": "clf_1", "target_set": "working", "output_prefix": "refined"}}
<<ENDIF:embeddings>>

explore_patterns  -- Dry-run 2-5 regex patterns WITHOUT creating sets.  Returns match counts and samples.  Compare ideas cheaply.
  {{"patterns": ["pat_a", "pat_b"], "target_set": "all"}}

set_operation  -- Combine two named sets (intersect, union, difference).
  {{"operation": "intersect", "set_a": "A", "set_b": "B", "result_name": "A_and_B"}}

read_docs  -- Read documents by set or IDs.  Use offset to sample different positions.  Set full_text=true for the complete document (no excerpting) -- free, zero LLM cost.  Use chunk_idx to page through a large document in contiguous chunks.
  {{"set_name": "my_set", "max_docs": 5, "query": "keywords"}}
  {{"set_name": "my_set", "max_docs": 3, "offset": -3}}
  {{"doc_ids": ["id"], "full_text": true}}
  {{"doc_ids": ["id"], "chunk_idx": 0}}
  # For huge docs, page through a contiguous slice:
  {{"doc_ids": ["id"], "start_char": 0, "max_chars": 50000}}

find_in_doc  -- Navigate inside ONE document by returning char offsets. Two modes: (1) exact match: provide pattern (+use_regex); (2) lexical passage ranking: provide query to surface high-overlap windows. In query mode, compare multiple matches using their previews and hit_tokens -- don't blindly take only the top-scored window if it doesn't contain the right section/table. Use with read_docs(start_char=..., max_chars=...) to jump directly in huge docs.
  {{"doc_id": "id", "pattern": "Balance Sheets", "max_matches": 10}}
  {{"doc_id": "id", "pattern": "cash and cash equivalents", "case_sensitive": false}}
  {{"doc_id": "id", "pattern": "Item\s+7\.?\s+Management", "use_regex": true}}
  {{"doc_id": "id", "query": "business segments net income 2022 Q2", "max_matches": 5}}

table_lookup  -- Find a specific row in aligned/table-like text and return the row + nearby header lines (often containing years/columns), plus parsed numbers. Use this to avoid mis-reading the wrong year/column in statements.
  {{"doc_id": "id", "row": "Total current liabilities", "max_matches": 5}}
  {{"doc_id": "id", "row": "Net property, plant, and equipment", "use_regex": true}}

count_set  -- Count docs in a set.  Shows verification snippets.
  {{"set_name": "my_set"}}

search_filenames  -- Find docs by filename pattern.  Supports target_set to restrict to an existing set.  Combine two filename searches with set_operation(intersect) for precise counts.
  {{"pattern": "value", "create_set": "matched"}}
  {{"pattern": "value", "target_set": "existing_set", "create_set": "refined"}}

create_regex  -- Register a regex or have the LLM create one from sample documents.
  {{"pattern": "your_regex"}}
  {{"condition": "what to detect", "sample_set_name": "working"}}

<<IF:embeddings>>
label_docs  -- LLM reads each doc and labels it 0/1 for conditions.
  {{"set_name": "working", "labels": ["condition"], "max_docs": 20}}

train_classifier  -- Train on embeddings + labels from label_docs.
  {{"label_batch_id": "label_0", "label_key": "condition"}}

cosine_search  -- Semantic search.  Creates a ranked top-k candidate set (default "cosine_top_k", or specify create_set); it is NOT an exhaustive semantic match set.  Use when keyword search returns 0 for a key entity.  Read from the created set, learn corpus structure, then build an exhaustive regex/index filter for the final answer.  Never use its cardinality as a corpus-level count or aggregate.
  {{"query": "description", "top_k": 10, "create_set": "my_cosine"}}
<<ENDIF:embeddings>>

<<IF:pattern_search>>
pattern_search  -- Query a pre-computed extraction index.  Use simple single-keyword params (e.g. "Smith" not "John.*Smith") for best recall.  Scope with target_set FIRST when the question constrains by year/category.  The result includes: (1) confirmed_matches -- keyword inside the captured structural region; (2) gap_clusters -- pre-computed structural templates showing where the keyword appears OUTSIDE the captured region, each with a template (keyword replaced by {{}}) and a count.  Compare each cluster template to the capture_sample and to your question -- read a few gap docs from the explore_set when a cluster may be relevant to the answer.
  {{"pattern": "name", "params": {{"key": "value"}}, "target_set": "some_set", "create_set": "result"}}
<<ENDIF:pattern_search>>

<<IF:agg_tools>>
extract_field  -- Extract ONE value from EACH doc using a regex capture group.  Stores results as a scalar value set (doc -> value).  Use group=1 for the first capture group or group="name" for a named group (?P<name>...).  as_type: "number" (default, parse as float), "text" (raw string), or "date" (parse date -> enables min/max for earliest/latest).  First read_docs to learn the exact value format, then write a regex that captures it reliably.
  {{"pattern": "your_regex_with_(capture)", "target_set": "filtered_set", "as_type": "number", "save_as": "my_values"}}
  {{"pattern": "(?:established|founded).*?(\w+ \d{{1,2}},? \d{{4}})", "target_set": "all", "as_type": "date", "save_as": "est_dates"}}

extract_list  -- Extract ALL matches of a regex from EACH doc as a list.  Unlike extract_field (one value per doc), this returns every match per doc.  Use for repeating items: names, line items, references, participants, etc.  Runs on raw document text (preserves newlines).  Follow with aggregate using list operations (unique_values, count_unique, value_counts, group_by_value).
  {{"pattern": "^([A-Z][a-z]+ [A-Z][a-z]+)", "target_set": "filtered_set", "save_as": "names"}}
  {{"pattern": "Item\s+(\d+)", "target_set": "all", "group": 1, "save_as": "item_numbers"}}

aggregate  -- Compute statistics over a named value set.  For scalar sets (from extract_field): sum, avg, min, max, count.  For min/max, returns the matching doc(s) and optionally creates a doc set for further inspection.  For list sets (from extract_list): unique_values (deduplicated list), count_unique (number of distinct items), value_counts (frequency table), flatten (all items), group_by_value (which docs share each value -- useful for cross-doc overlap questions), per_doc_count (total items per doc -> scalar value set), per_doc_count_unique (unique items per doc -> scalar value set).  per_doc_count_unique is essential for "average unique X per doc" questions: first extract_list, then aggregate(per_doc_count_unique, create_set="counts"), then aggregate(value_set="counts", operation="avg").
  {{"value_set": "my_values", "operation": "avg"}}
  {{"value_set": "my_values", "operation": "min", "create_set": "lowest_docs"}}
  {{"value_set": "names", "operation": "unique_values"}}
  {{"value_set": "names", "operation": "per_doc_count_unique", "create_set": "name_counts"}}

extract_from_filename  -- Extract metadata from each document's FILENAME using a regex capture group.  Unlike extract_field (which runs on document text), this runs on the filename string.  Use for metadata encoded in filenames: company tickers, dates, quarters, categories.  Stores results as a scalar value set (doc -> captured string).
  {{"pattern": "^([A-Z]+)_", "target_set": "all", "save_as": "company_ticker"}}
  {{"pattern": "(\d{{4}}-Q\d)", "target_set": "all", "save_as": "quarter"}}

filter_values  -- Filter docs by comparing their extracted value to a threshold.  Takes a value set from extract_field, an operator (<, <=, >, >=, ==, !=), and a threshold.  Creates a new doc set containing only the docs that pass.  For dates, the threshold can be a date string (e.g. "2020-01-01").
  {{"value_set": "ratings", "operator": "<", "threshold": 8.23, "create_set": "low_rated"}}
  {{"value_set": "est_dates", "operator": ">=", "threshold": "2020-01-01", "create_set": "recent"}}
<<ENDIF:agg_tools>>

calculate  -- Evaluate math expressions safely.  Supports +, -, *, /, //, %, ** (including fractional exponents like x**0.5), parentheses, and functions: sqrt, cbrt, log, log2, log10, ln, exp, abs, round, ceil, floor, pow, min, max, sum, factorial, gcd.  Constants: pi, e.  Use for ANY arithmetic -- never do mental math.  For multi-step calculations, pass a list of named expressions; each step can reference earlier step names as variables.
  {{"expr": "1234.5 / 67.8"}}
  {{"expr": "(revenue - cost) / revenue * 100", "variables": {{"revenue": 5400, "cost": 3200}}}}
  {{"expressions": [{{"name": "margin", "expr": "revenue - cost"}}, {{"name": "pct", "expr": "margin / revenue * 100"}}], "variables": {{"revenue": 5400, "cost": 3200}}}}

list_sets<<IF:embeddings>> / list_classifiers<<ENDIF:embeddings>>  -- Inspect current state.

=======================================================================
GUARDRAILS
=======================================================================

* "Contains X" != "Is about X."  Always refine before answering.
* Validate from different positions -- beginning, middle, and end.
* Watch alternation: "A|B:.*X" means "A" OR "B:.*X".  Use "(?:A|B:).*X" for what you probably intend.
* Different format != different answer.  UNION format variants.
* Record findings in your thought -- observations get redacted.
* SCOPE FIDELITY.  Do not narrow beyond what the question asks.  If the question uses a general term, your answer must cover the full category, not a single sub-type.
* BE VERY SKEPTICAL OF ZERO RESULTS.  0 matches usually means your pattern is wrong, not the data.  Try: shorter substrings, different spellings, broader regex<<IF:embeddings>>, cosine_search with a natural language description<<ENDIF:embeddings>>.  Exhaust at least 3 different strategies before concluding information is absent.
* NEVER STOP AT AN INTERMEDIATE FINDING.  If you resolved "A is associated with B" but the question asks about a property OF B, you must keep searching.  Re-read your sub-question decomposition.
* INSUFFICIENT_EVIDENCE is a claim that requires evidence.  You may only conclude "insufficient evidence" if you searched thoroughly and can state what you searched for and why nothing matched.  If you have partial findings, give your best answer and state what is uncertain.
* COUNTING: count_set gives an EXACT doc count.  Answer "how many" with a single integer from count_set.  NEVER say "at least N" or "approximately N" -- hedging is always wrong.  If uncertain about filter quality, refine the set and re-count instead of hedging.
<<IF:pattern_search>>
* CONFIRMED vs. GAP: pattern_search reports confirmed_matches (keyword inside the captured structural region) and gap_clusters (keyword OUTSIDE it, in a different structural position).  The captured region defines what the pattern was designed to extract.  A gap cluster with a DIFFERENT surrounding-text pattern than the capture_sample means the keyword serves a different function there -- do NOT count those docs.  Only read gap docs (from the explore_set) when a gap cluster template is genuinely ambiguous and you cannot determine from the template alone whether it satisfies the question.  The confirmed count is high-confidence; gap additions need clear evidence of equivalence.
<<ENDIF:pattern_search>>
* CORPUS ONLY.  Your pre-trained knowledge may conflict with this corpus.  An abbreviation, name, or term may have a corpus-specific meaning that differs from the real world.  Always derive meaning FROM the documents, never from memory.
* read_docs FOCUS.  When you need a specific section of a large document (e.g., a clause, a date, a specific term), pass query="the specific terms" to focus excerpts on what you need.  If excerpts still miss the section, page contiguously with chunk_idx (coarse) or start_char/max_chars (precise).  Use find_in_doc to get offsets for start_char: use pattern/use_regex when you know the exact phrase/section title; if you only have a natural-language description, use find_in_doc(query=...) to surface likely windows (compare previews + hit_tokens), then page contiguously with start_char/max_chars.  Without query, excerpts show the filter-match context.
* {max_steps} tool calls.  Typical: 5-15 steps.

=======================================================================
SELF-CHECK BEFORE ANSWERING
=======================================================================

[ ] Did I resolve ALL sub-questions, or am I answering an intermediate hop?
[ ] Is my answer based on text I actually read (read_docs), not just keyword counts or snippet previews?
[ ] For counting: is my count a single integer from count_set -- not hedged with "at least" or "approximately"?
<<IF:agg_tools>>
[ ] For aggregation (avg/sum/min/max): did I use extract_field + aggregate -- not manual addition or estimation?
[ ] For list/enumeration questions: did I use extract_list + aggregate(unique_values/count_unique) -- not manual counting?
<<ENDIF:agg_tools>>
[ ] For any arithmetic (division, percentages, ratios, differences): did I use calculate -- not mental math?
[ ] Percent vs ratio: ONLY multiply by 100 or add a % sign if the question explicitly asks for a percentage / percent / %. Otherwise, report ratios as decimals (e.g., 0.83), not percent (83%).
[ ] Does my answer satisfy EVERY constraint in the original question?
[ ] Am I using the CORPUS meaning of terms, not my own world knowledge?

=======================================================================
RESPONSE FORMAT
=======================================================================

Each turn: exactly ONE JSON object, nothing else.

Tool call:
  {{"thought": "what I plan to do and why", "tool": "name", "args": {{...}}}}

Final answer:
  {{"thought": "summary of evidence", "answer": "precise answer"}}
\end{lstlisting}

\section{Context Management Details}
\label{app:context}

\paragraph{Question landscape.}
Before the agent's first turn, we extract all maximal alphanumeric tokens of length $\geq 4$ from the question ("keywords").  For the first five keywords (in order of appearance), the system scans the full corpus via case-insensitive substring match and records: (i) the number of documents whose body text contains the keyword, and (ii) the number of filenames containing it.  For four keywords that match at least $1$ but less than all of the documents (sorted by ascending match count), we additionally compute and report pairwise intersection sizes, i.e., the number of documents containing both keywords simultaneously.  A 600-character snippet from the top keyword's first match is appended, showing the agent one formatted example.  The entire landscape is prepended to the user message.

\paragraph{Tokenization and matching rules (for disambiguation).}
The code uses similar-looking but distinct lexical operations in different places; we list them explicitly to avoid ambiguity.
\begin{itemize}\setlength{\itemsep}{1pt}
\item \textbf{Question landscape tokens (``keywords'').} Tokens are extracted from the question via \verb|[^\W_]{4,}|.  A token is counted as present in a document if it matches by case-insensitive substring inclusion (\texttt{tok.lower() in text.lower()}).  These counts provide a cheap prior only.
\item \textbf{Anchor tokens (``entities'').} The system sorts a subset the keywords with \(0 < \mathrm{count}(tok) < N\), preferring tokens that appear capitalized in the original question and then by increasing document frequency.  These anchor tokens are used only for entity-aware diagnostics (scope checks; negative-entity checks), not as a learned NER component.
\item \textbf{\texttt{apply\_filter} decomposition tokens.} For diagnostics, tokens are extracted from the \emph{regex string} with the same tokenizer \verb|[^\W_]{4,}|.  Each token is then probed as a single-token case-insensitive regex over normalized text to estimate per-token selectivity and token intersections.
\item \textbf{Extraction-index value tokens (gap clusters).} During indexing, tokens are extracted from captured values via \verb|[^\W\d_]{3,}| (letter-only, length \(\ge 3\)), lowercased, and used to pre-compute \texttt{keyword\_gaps}: cases where a value token appears in a document but outside the captured region.  \texttt{pattern\_search} surfaces these as gap clusters (surrounding-text templates with the token replaced by \texttt{\{\}}).
\end{itemize}

\paragraph{Context management overview.}
The agent's context window must accommodate the system prompt, the question, the question landscape, and a growing sequence of tool observations and agent responses.  Because individual documents can be hundreds of thousands of characters and trajectories can span 25 steps, two mechanisms prevent overflow while preserving the set-algebraic state that the agent needs for planning.

\paragraph{Per-tool output budgets.}
Every tool observation is hard-capped at $B_{\mathrm{tool}}$ characters (set to 200{,}000 for models with $\geq$500K context, proportionally less for smaller models).  For \texttt{read\_docs} specifically, the budget is further divided evenly among the requested documents: when reading $k$ documents, each gets $\lfloor B_{\mathrm{call}} / k \rfloor$ characters, where $B_{\mathrm{call}} = \min(C_{\max}/6, B_{\mathrm{tool}})$.  Documents exceeding their per-doc budget are presented via \emph{keyword-guided excerpting}: the system identifies text windows around query-relevant terms and concatenates only those passages.  For long-document navigation without relying on excerpting, the agent can also page contiguously (chunked reads) or request explicit \texttt{start\_char}/\texttt{max\_chars} slices.

\paragraph{Observation redaction.}
After the agent responds to an observation (i.e., the observation is ``consumed"), the full text is replaced by a structural summary.  The summary retains a whitelisted set of JSON keys:
\begin{itemize}\setlength{\itemsep}{0pt}
\item \emph{Set identity}: \texttt{positive\_set}, \texttt{negative\_set}, \texttt{result\_set}, \texttt{created\_set}.
\item \emph{Cardinalities}: \texttt{positive\_count}, \texttt{negative\_count}, \texttt{total\_filtered}, \texttt{count}.
\item \emph{Diagnostics}: \texttt{keyword\_decomposition} (per-term match counts), \texttt{entity\_coverage}, \texttt{selectivity\_gap}, \texttt{zero\_result\_hint}, \texttt{scope\_check}, \texttt{negative\_entity\_check}.
\item \emph{Extraction state}: \texttt{value\_set}, \texttt{extracted}, \texttt{scanned}, \texttt{n\_values}, \texttt{matching\_docs}.
\end{itemize}

Everything else (raw document text, full match listings) is discarded.  The agent's own \texttt{thought} field (which the system prompt instructs it to use as a scratchpad) preserves factual findings across turns. The external workspace preserves executable retrieval state, while the LLM context carries planning and textual reasoning; factual observations become typed objects when subsequent operations need to execute over them.

\paragraph{Last-resort compression.}
If the total conversation still exceeds the budget $B = 0.6 \times C_{\max}$ characters after redaction, messages in the middle of the conversation (excluding the system prompt, the initial question, and the four most recent messages) are further truncated to their first 150 characters.  If still over budget, middle messages are dropped entirely.

The net effect is that the \emph{workspace state} (the names, sizes, provenance, and diagnostic metadata of all sets) is always visible to the agent, while raw document content is available only for the most recent observation and must be captured in the agent's thought before it becomes redacted.

\section{Trace Example}
\label{app:traces}

The following representative trace is copied verbatim from the logs. It
illustrates how persistent sets and diagnostic feedback guide the agent's
reasoning.

\subsection{Multi-Hop with Diagnostic Recovery: 2WikiMultiHopQA}
\label{app:trace-2wiki}

This trace illustrates a two-hop question solved by constructing and comparing named document sets. The agent starts with a broad lexical filter, then refines by intersecting constraints (song title, composer mention), and uses diagnostics such as \texttt{scope\_check} and \texttt{negative\_entity\_check} plus auto-sampled snippets to confirm the correct entity before extracting the birthplace.

\begin{lstlisting}
Question: Where was the composer of song Back In The U.S.A. born?
Golden answer: St. Louis, Missouri
Agent answer: St. Louis, Missouri

Step 0: apply_filter({
  "filter_type": "regex",
  "pattern": "composer",
  "target_set": "all",
  "output_prefix": "has_composer"
})
  Agent thought: The question is: Where was the composer of the song 'Back In The U.S.A.' born? This is a two-hop question: (1) Who is the composer of 'Back In The U.S.A.'? (2) Where was that composer born? I will first search for documents mentioning both 'composer' and 'Back' to identify the composer, then search for the composer's birthplace.
  Observation: {
  "positive_set": "has_composer_pos",
  "positive_count": 116,
  "negative_set": "has_composer_neg",
  "negative_count": 6002,
  "total_filtered": 6118,
  "sample_positive_docs": [
    {
      "doc_id": "A_Time_of_Destiny__1b00ec88a396",
      "filename": "A_Time_of_Destiny.txt"
    },
    {
      "doc_id": "Abd\u00fclaziz__6cfec66ec8bc",
      "filename": "Abd\u00fclaziz.txt"
    },
    {
      "doc_id": "Keith_Emerson__f2d70a357c06",
      "filename": "Keith_Emerson.txt"
    },
    {
      "doc_id": "Kurt_Schr\u00f6der__4830d105778e",
      "filename": "Kurt_Schr\u00f6der.txt"
    },
    {
      "doc_id": "Youve_Got_Possibilities__e9a4993a16da",
      "filename": "Youve_Got_Possibilities.txt"
    },
    {
      "doc_id": "\u00c1ngel_Recasens__6497d74babaa",
      "filename": "\u00c1ngel_Recasens.txt"
    }
  ],
  "sample_negative_docs": [
    {
      "doc_id": "11_Harrowhouse__08cc6c95c2f9",
      "filename": "11_Harrowhouse.txt"
    },
    {
      "doc_id": "1971_Copa_Libertadores__8369d1b96cbe",
      "filename": "1971_Copa_Libertadores.txt"
    },
    {
      "doc_id": "Khalid_Abdel_Nasser__b9f61f804f4b",
      "filename": "Khalid_Abdel_Nasser.txt"
    },
    {
      "doc_id": "Khalid_al-Habib__ca209876aaad",
      "filename": "Khalid_al-Habib.txt"
    },
    {
      "doc_id": "\u021aerova__10c94b229856",
      "filename": "\u021aerova.txt"
    },
    {
      "doc_id": "\u02bbElisiva_Fusipala_Tauki\u02bbonetuku__3e99f95742e5",
      "filename": "\u02bbElisiva_Fusipala_Tauki\u02bbonetuku.txt"
    }
  ],
  "scope_check": "apply_filter searches DOCUMENT TEXT. 116 docs matched in text, but 4 filenames match the same pattern. If you intended to filter by filename/date, use search_filenames instead.",
  "negative_entity_check": "139/6002 negatives contain 'Back' (keyword only)."
}
[Auto-sampled snippets (positive) -- 3 of 116 docs, explore further if needed]
  [A_Time_of_Destiny.txt] A Time of Destiny

A Time of Destiny is a 1988 American drama film directed by Gregory Nava and written by Nava and Anna Thomas. The story is based on the opera" La forza del destino" by Giuseppe Verdi. The motion picture was executive produced by Shep Gordon and Carolyn Pfeiffer. It features original music by veteran composer Ennio Morricone. Set during World War II in Italy and San Diego, the film tells of two friends who become enemies during the war.
  [Kurt_Schroder.txt] Kurt Schroder

Kurt Schroder( 1888-1962) was a German composer. Schroder composed a number of film scores. During the 1930s he worked in Britain for Alexander Korda's London Film Productions, and scored the company's breakthrough hit" The Private Life of Henry VIII" in 1933.
  [Angel_Recasens.txt] Angel Recasens

Angel Recasens( 4 March 1938 in Cambrils - 2 August 2007) was a Catalan organist, teacher, composer and musicologist best known as a choral conductor. From 1975 to 1986, he was director of the Coro de Sant Esteve of Vila- seca. There, he performed music from the romantics Schumann and Mendelssohn, to the contemporary music of Ligeti and Schnittke. He also directed the Coral Verge del Cami in his home town Cambrils. In his later years, he dedicated himself to reviving lost Spanish choral works from the 16th to 18th centuries; first with la Capilla Principe de Viana, then with the expanded La Grande Chapelle and his own label Lauda Records. Both the ensemble and the label were continued by his son Albert Recasens.
[Auto-sampled snippets (negative) -- 3 of 6002 docs, explore further if needed]
  [11_Harrowhouse.txt] 11 Harrowhouse

11 Harrowhouse is a 1974 British DeLuxe Color film directed by Aram Avakian and starring Charles Grodin, Candice Bergen, James Mason, Trevor Howard and John Gielgud in Panavision. It was adapted by Charles Grodin based upon the novel by Gerald A. Browne with the screenplay by Jeffrey Bloom.
  [Khalid_al-Habib.txt] Khalid al-Habib

Khalid al- Habib also believed to use the nom de guerre Khalid al Harbi is an Al- Qaeda leader who has overseen Al- Qaeda operations in Afghanistan. In 2005 it was reported that al- Habib died in a US missile strike on the Pakistani village of Damadola though it is unclear whether this is the case as Pakistani officials have claimed that no Al- Qaeda leaders died in the attack. al- Habib's nationality is not clear with sources claiming he may be Egyptian or Moroccan.
  ['Elisiva_Fusipala_Tauki'onetuku.txt] 'Elisiva Fusipala Tauki'onetuku

'Elisiva Fusipala Tauki'onetuku (18 May 1850 - September 1889) was the mother of King George Tupou II.

=== Automatically computed ===
Filename search counts:
  search_filenames("composer") -> 4 docs

Step 1: apply_filter({
  "filter_type": "regex",
  "pattern": "Back",
  "target_set": "has_composer_pos",
  "output_prefix": "composer_and_Back"
})
  Agent thought: Now I need to narrow to docs that mention both 'composer' and 'Back' to find the composer of 'Back In The U.S.A.'. This will help identify the composer before searching for their birthplace.
  Observation: {
  "positive_set": "composer_and_Back_pos",
  "positive_count": 7,
  "negative_set": "composer_and_Back_neg",
  "negative_count": 109,
  "total_filtered": 116,
  "sample_positive_docs": [
    {
      "doc_id": "Deva_composer__371eabd0fe49",
      "filename": "Deva_composer.txt"
    },
    {
      "doc_id": "Johann_Sebastian_Bach__d44cb4c567c7",
      "filename": "Johann_Sebastian_Bach.txt"
    },
    {
      "doc_id": "Mario_Bonnard__be124b2927fc",
      "filename": "Mario_Bonnard.txt"
    },
    {
      "doc_id": "Michael_Corcoran_musician__f988e03f8c59",
      "filename": "Michael_Corcoran_musician.txt"
    },
    {
      "doc_id": "Raffy_Shart__4b2c2178b7ca",
      "filename": "Raffy_Shart.txt"
    },
    {
      "doc_id": "Youve_Got_Possibilities__e9a4993a16da",
      "filename": "Youve_Got_Possibilities.txt"
    }
  ],
  "sample_negative_docs": [
    {
      "doc_id": "A_Time_of_Destiny__1b00ec88a396",
      "filename": "A_Time_of_Destiny.txt"
    },
    {
      "doc_id": "Abd\u00fclaziz__6cfec66ec8bc",
      "filename": "Abd\u00fclaziz.txt"
    },
    {
      "doc_id": "Karel_Zich__efc220f3e985",
      "filename": "Karel_Zich.txt"
    },
    {
      "doc_id": "Karl_Wilhelm_conductor__fdea54ab7eab",
      "filename": "Karl_Wilhelm_conductor.txt"
    },
    {
      "doc_id": "William_Axt__f2e5b44b3b94",
      "filename": "William_Axt.txt"
    },
    {
      "doc_id": "\u00c1ngel_Recasens__6497d74babaa",
      "filename": "\u00c1ngel_Recasens.txt"
    }
  ],
  "scope_check": "apply_filter searches DOCUMENT TEXT. 7 docs matched in text, but 0 filenames match the same pattern. If you intended to filter by filename/date, use search_filenames instead."
}
[Auto-sampled snippets (positive) -- 3 of 7 docs, explore further if needed]
  [Deva_composer.txt] Deva (composer)

Devanesan Chokkalingam, popularly known as Deva, is an Indian film composer and singer. He has composed songs and provided background music for Tamil, Telugu, Malayalam and Kannada films in a career spanning about 20 years. Many know his gaana songs, written mostly using Madras Tamil. He is known as the "Father of Gaana Genre" in the Tamil film industry. Deva has composed music for many films. He debuted as a film music director in the film "Manasukkeththa Maharaasa" in 1989. In the intervening years he has composed music for a total of more than 400 films.
  [Michael_Corcoran_musician.txt] Michael Corcoran (musician)

Michael Thomas Corcoran( born December 9, 1972), known professionally as Backhouse Mike or Ken Lofkoll, is an American musician, singer- songwriter, composer, sound designer, musical arranger, orchestrator, and record producer.
  [Youve_Got_Possibilities.txt] ...e pry you from your shell ... You've got possibilities ... you do n't even know you've got"). " It's a Bird ... It's a Plane ... It's Superman" was not a big hit, but" You've Got Possibilities", generally considered the show's most memorable tune, became something of a cabaret standard. Peggy Lee recorded the song on her 1966 album" Big$ pender" and released it as the B-side of the single" Come Back To Me". Joanie Sommers released" You've Got Possibilities" as the B-side of her single" Never Throw Your Dreams Away", also in 1966, while Carol Ventura released it also in 1966 as an A- side single. Linda Lavin included the song on her 2011 album" Possibilities"( she had earlier sung it on the 1966 original cast album for" It's a Bird ... It's a Plane ... It's Superman"). Matt Monro's version...
[Auto-sampled snippets (negative) -- 3 of 109 docs, explore further if needed]
  [A_Time_of_Destiny.txt] A Time of Destiny

A Time of Destiny is a 1988 American drama film directed by Gregory Nava and written by Nava and Anna Thomas. The story is based on the opera" La forza del destino" by Giuseppe Verdi. The motion picture was executive produced by Shep Gordon and Carolyn Pfeiffer. It features original music by veteran composer Ennio Morricone. Set during World War II in Italy and San Diego, the film tells of two friends who become enemies during the war.
  [Karl_Wilhelm_conductor.txt] Karl Wilhelm (conductor)

Karl Wilhelm, also Carl Wilhelm( 5 September 1815, Schmalkalden - 26 August 1873, Schmalkalden) was a German choral director. He is best known as the composer of the music of the song " Die Wacht am Rhein. "
  [Angel_Recasens.txt] Angel Recasens

Angel Recasens( 4 March 1938 in Cambrils - 2 August 2007) was a Catalan organist, teacher, composer and musicologist best known as a choral conductor. From 1975 to 1986, he was director of the Coro de Sant Esteve of Vila- seca. There, he performed music from the romantics Schumann and Mendelssohn, to the contemporary music of Ligeti and Schnittke. He also directed the Coral Verge del Cami in his home town Cambrils. In his later years, he dedicated himself to reviving lost Spanish choral works from the 16th to 18th centuries; first with la Capilla Principe de Viana, then with the expanded La Grande Chapelle and his own label Lauda Records. Both the ensemble and the label were continued by his son Albert Recasens.
[Context audit -- 'Back' in 5 of 7 positive docs:]
  ...ger. He has composed songs and provided background music for Tamil, Telugu, Malayala...
  ...many in films directed by Luis Trenker. Back in Italy in 1932, he became a prolific...
  ...ember 9, 1972), known professionally as Backhouse Mike or Ken Lofkoll, is an America...
  ...king at the Uberbrettl cabaret. He went back to Vienna and began writing operettas,...
  ...ed it as the B-side of the single" Come Back To Me". Joanie Sommers released" You've...

=== Automatically computed ===
Filename search counts:
  search_filenames("Back") -> 11 docs
Set compositions -- has_composer_pos (116) x composer_and_Back_pos (7):
  intersect -> 7
  union     -> 116
  has_composer_pos \ composer_and_Back_pos -> 109
  composer_and_Back_pos \ has_composer_pos -> 0

Step 2: apply_filter({
  "filter_type": "regex",
  "pattern": "Back In The U\\.S\\.A\\.",
  "target_set": "all",
  "output_prefix": "has_BackInUSA"
})
  Agent thought: None of the 7 docs in composer_and_Back_pos appear to be about the song 'Back In The U.S.A.'--the matches are for 'Backhouse Mike', 'Come Back To Me', and other unrelated uses of 'Back'. I need to try a broader search for the song title 'Back In The U.S.A.' to identify the composer directly. I will search for 'Back In The U.S.A.' in all docs.
  Observation: {
  "positive_set": "has_BackInUSA_pos",
  "positive_count": 1,
  "negative_set": "has_BackInUSA_neg",
  "negative_count": 6117,
  "total_filtered": 6118,
  "sample_positive_docs": [
    {
      "doc_id": "Back_in_the_USA__a505370ce874",
      "filename": "Back_in_the_USA.txt"
    }
  ],
  "sample_negative_docs": [
    {
      "doc_id": "11_Harrowhouse__08cc6c95c2f9",
      "filename": "11_Harrowhouse.txt"
    },
    {
      "doc_id": "1971_Copa_Libertadores__8369d1b96cbe",
      "filename": "1971_Copa_Libertadores.txt"
    },
    {
      "doc_id": "Khalid_Abdel_Nasser__b9f61f804f4b",
      "filename": "Khalid_Abdel_Nasser.txt"
    },
    {
      "doc_id": "Khalid_al-Habib__ca209876aaad",
      "filename": "Khalid_al-Habib.txt"
    },
    {
      "doc_id": "\u021aerova__10c94b229856",
      "filename": "\u021aerova.txt"
    },
    {
      "doc_id": "\u02bbElisiva_Fusipala_Tauki\u02bbonetuku__3e99f95742e5",
      "filename": "\u02bbElisiva_Fusipala_Tauki\u02bbonetuku.txt"
    }
  ],
  "scope_check": "apply_filter searches DOCUMENT TEXT. 1 docs matched in text, but 0 filenames match the same pattern. If you intended to filter by filename/date, use search_filenames instead.",
  "negative_entity_check": "145/6117 negatives contain 'Back' (keyword only)."
}
[Auto-sampled snippets (positive) -- 3 of 1 docs, explore further if needed]
  [Back_in_the_USA.txt] Back in the U.S.A.

"Back in the U.S.A." is a song written by Chuck Berry that was released in 1959 and was a top 40 hit. A cover version in 1978 by Linda Ronstadt was also a hit.
[Auto-sampled snippets (negative) -- 3 of 6117 docs, explore further if needed]
  [11_Harrowhouse.txt] 11 Harrowhouse

11 Harrowhouse is a 1974 British DeLuxe Color film directed by Aram Avakian and starring Charles Grodin, Candice Bergen, James Mason, Trevor Howard and John Gielgud in Panavision. It was adapted by Charles Grodin based upon the novel by Gerald A. Browne with the screenplay by Jeffrey Bloom.
  [Khalid_al-Habib.txt] Khalid al-Habib

Khalid al- Habib also believed to use the nom de guerre Khalid al Harbi is an Al- Qaeda leader who has overseen Al- Qaeda operations in Afghanistan. In 2005 it was reported that al- Habib died in a US missile strike on the Pakistani village of Damadola though it is unclear whether this is the case as Pakistani officials have claimed that no Al- Qaeda leaders died in the attack. al- Habib's nationality is not clear with sources claiming he may be Egyptian or Moroccan.
  ['Elisiva_Fusipala_Tauki'onetuku.txt] 'Elisiva Fusipala Tauki'onetuku

'Elisiva Fusipala Tauki'onetuku (18 May 1850 - September 1889) was the mother of King George Tupou II.

=== Automatically computed ===
Filename search counts:
  search_filenames("Back") -> 11 docs
Set compositions -- composer_and_Back_pos (7) x has_BackInUSA_pos (1):
  intersect -> 0
  union     -> 8
  composer_and_Back_pos \ has_BackInUSA_pos -> 7
  has_BackInUSA_pos \ composer_and_Back_pos -> 1

Step 3: apply_filter({
  "filter_type": "regex",
  "pattern": "Chuck Berry",
  "target_set": "all",
  "output_prefix": "has_ChuckBerry"
})
  Agent thought: The document 'Back_in_the_USA.txt' states: 'Back in the U.S.A.' is a song written by Chuck Berry. This answers sub-question 1: the composer is Chuck Berry. Now, I need to find where Chuck Berry was born. I will search for 'Chuck Berry' and 'born' to find his birthplace.
  Observation: {
  "positive_set": "has_ChuckBerry_pos",
  "positive_count": 3,
  "negative_set": "has_ChuckBerry_neg",
  "negative_count": 6115,
  "total_filtered": 6118,
  "sample_positive_docs": [
    {
      "doc_id": "Alan_Reeves_composer__89c0d8e6240e",
      "filename": "Alan_Reeves_composer.txt"
    },
    {
      "doc_id": "Back_in_the_USA__a505370ce874",
      "filename": "Back_in_the_USA.txt"
    },
    {
      "doc_id": "Chuck_Berry__ff7ec8ef49a2",
      "filename": "Chuck_Berry.txt"
    }
  ],
  "sample_negative_docs": [
    {
      "doc_id": "11_Harrowhouse__08cc6c95c2f9",
      "filename": "11_Harrowhouse.txt"
    },
    {
      "doc_id": "1971_Copa_Libertadores__8369d1b96cbe",
      "filename": "1971_Copa_Libertadores.txt"
    },
    {
      "doc_id": "Khalid_al-Habib__ca209876aaad",
      "filename": "Khalid_al-Habib.txt"
    },
    {
      "doc_id": "Khalid_bin_Abdullah_bin_Abdulaziz_Al_Saud__
      b63dd78ee252",
      "filename": "Khalid_bin_Abdullah_bin_Abdulaziz_Al_Saud.txt"
    },
    {
      "doc_id": "\u021aerova__10c94b229856",
      "filename": "\u021aerova.txt"
    },
    {
      "doc_id": "\u02bbElisiva_Fusipala_Tauki\u02bbonetuku__3e99f95742e5",
      "filename": "\u02bbElisiva_Fusipala_Tauki\u02bbonetuku.txt"
    }
  ],
  "scope_check": "apply_filter searches DOCUMENT TEXT. 3 docs matched in text, but 0 filenames match the same pattern. If you intended to filter by filename/date, use search_filenames instead.",
  "keyword_decomposition": [
    {
      "keyword": "Chuck",
      "matches": 10,
      "pct": 0.2
    },
    {
      "keyword": "Berry",
      "matches": 21,
      "pct": 0.3
    }
  ],
  "selectivity_note": "Combined pattern matched 3; individual keywords match 10\u201321 each. Consider starting with one keyword.",
  "keyword_intersection": {
    "keywords": [
      "Chuck",
      "Berry"
    ],
    "both_match": 3
  },
  "negative_entity_check": "144/6115 negatives contain 'Back' (keyword only)."
}
[Auto-sampled snippets (positive) -- 3 of 3 docs, explore further if needed]
  [Alan_Reeves_composer.txt] Alan Reeves (composer)

Alan David Reeves is a British film composer, music producer, and Hammond B3 virtuoso. In the course of his career he has received 35 international awards, including a Goldene Schallplatte. He became known for his work with the bands The Showtimers and Clinic as well as for the music for the films To Walk with Lions,, and Kill Bill Vol. 2. He has appeared played or recorded with/ for among others, the Rolling Stones, Jimi Hendrix, Chuck Berry and David Bowie and David Gilmore
  [Back_in_the_USA.txt] Back in the U.S.A.

"Back in the U.S.A." is a song written by Chuck Berry that was released in 1959 and was a top 40 hit. A cover version in 1978 by Linda Ronstadt was also a hit.
  [Chuck_Berry.txt] Chuck Berry

Charles Edward Anderson Berry (October 18, 1926 - March 18, 2017) was an American singer and songwriter, and one of the pioneers of rock and roll music. Nicknamed the "Father of Rock and Roll", Berry refined and developed rhythm and blues into the major elements that made rock and roll distinctive with songs such as "Maybellene" (1955), " Roll Over Beethoven" (1956), "Rock and Roll Music" (1957) and "Johnny B. Goode" (1958). Writing lyrics that focused on teen life and consumerism, and developing a music style that included guitar solos and showmanship, Berry was a major influence on subsequent rock music. Born into a middle-class African-American family in St. Louis, Missouri, Berry had an interest in music from an early age and gave his first public performance at Sumner Hig...
[Auto-sampled snippets (negative) -- 3 of 6115 docs, explore further if needed]
  [11_Harrowhouse.txt] 11 Harrowhouse

11 Harrowhouse is a 1974 British DeLuxe Color film directed by Aram Avakian and starring Charles Grodin, Candice Bergen, James Mason, Trevor Howard and John Gielgud in Panavision. It was adapted by Charles Grodin based upon the novel by Gerald A. Browne with the screenplay by Jeffrey Bloom.
  [Khalid_bin_Abdullah_bin_Abdulaziz_Al_Saud.txt] Khalid bin Abdullah bin Abdulaziz Al Saud

Khalid bin Abdullah( born 1950, Riyadh) is the eldest child of King Abdullah of Saudi Arabia. He is a member of the Allegiance Council.
  ['Elisiva_Fusipala_Tauki'onetuku.txt] 'Elisiva Fusipala Tauki'onetuku

'Elisiva Fusipala Tauki'onetuku (18 May 1850 - September 1889) was the mother of King George Tupou II.
[Context audit -- 'Back' in 2 of 3 positive docs:]
  Back in the U.S.A.  "Back in the U.S.A." is...
  ...ormer, playing his past hits with local backup bands of variable quality. However, i...

=== Automatically computed ===
Filename search counts:
  search_filenames("Chuck") -> 2 docs
  search_filenames("Berry") -> 3 docs
Set compositions -- has_BackInUSA_pos (1) x has_ChuckBerry_pos (3):
  intersect -> 1
  union     -> 3
  has_BackInUSA_pos \ has_ChuckBerry_pos -> 0
  has_ChuckBerry_pos \ has_BackInUSA_pos -> 2
\end{lstlisting}

\subsection{Zero-Result Diagnostics and Recovery: 2WikiMultiHopQA}
\label{app:trace-recovery}

Here the agent's initial detention-related regexes yield zero results within its already-narrowed working set. The \texttt{zero\_result\_hint} diagnostic prompts it to try broader lexical variants (e.g., \texttt{detain} $\rightarrow$ \texttt{arrest} $\rightarrow$ \texttt{prison|jail}), and the agent recovers without widening the search back to the full corpus.

\begin{lstlisting}
Question: Where was the performer of song B Boy (Song) detained?
Golden answer: State Correctional Institution - Chester
Agent answer: State Correctional Institution - Chester in Chester, Pennsylvania.

Step 0: apply_filter({
  "filter_type": "regex",
  "pattern": "B Boy",
  "target_set": "all",
  "output_prefix": "has_b_boy"
})
  Agent thought: This is a multi-hop question: (1) Identify the performer of 'B Boy (Song)'; (2) Determine where that performer was detained. I will start by searching for 'B Boy' to identify the performer, then search for detention information about that person.
  Observation: {
  "positive_set": "has_b_boy_pos",
  "positive_count": 1,
  "negative_set": "has_b_boy_neg",
  "negative_count": 6117,
  "total_filtered": 6118,
  "sample_positive_docs": [
    {
      "doc_id": "B_Boy_song__638a840ec141",
      "filename": "B_Boy_song.txt"
    }
  ],
  "sample_negative_docs": [
    {
      "doc_id": "11_Harrowhouse__08cc6c95c2f9",
      "filename": "11_Harrowhouse.txt"
    },
    {
      "doc_id": "1971_Copa_Libertadores__8369d1b96cbe",
      "filename": "1971_Copa_Libertadores.txt"
    },
    {
      "doc_id": "Khalid_Abdel_Nasser__b9f61f804f4b",
      "filename": "Khalid_Abdel_Nasser.txt"
    },
    {
      "doc_id": "Khalid_al-Habib__ca209876aaad",
      "filename": "Khalid_al-Habib.txt"
    },
    {
      "doc_id": "\u021aerova__10c94b229856",
      "filename": "\u021aerova.txt"
    },
    {
      "doc_id": "\u02bbElisiva_Fusipala_Tauki\u02bbonetuku__3e99f95742e5",
      "filename": "\u02bbElisiva_Fusipala_Tauki\u02bbonetuku.txt"
    }
  ],
  "scope_check": "apply_filter searches DOCUMENT TEXT. 1 docs matched in text, but 0 filenames match the same pattern. If you intended to filter by filename/date, use search_filenames instead.",
  "negative_entity_check": "279/6117 negatives contain 'Where' (keyword only)."
}
[Auto-sampled snippets (positive) -- 3 of 1 docs, explore further if needed]
  [B_Boy_song.txt] B Boy (song)

"B Boy" is a song by American hip hop recording artist Meek Mill. It was released as a third single from "Dreams Worth More Than Money" on January 3, 2015, by Maybach Music Group and Atlantic Records. The song, which was produced by Sap, features guest appearances from Big Sean and ASAP Ferg.
[Auto-sampled snippets (negative) -- 3 of 6117 docs, explore further if needed]
  [11_Harrowhouse.txt] 11 Harrowhouse

11 Harrowhouse is a 1974 British DeLuxe Color film directed by Aram Avakian and starring Charles Grodin, Candice Bergen, James Mason, Trevor Howard and John Gielgud in Panavision. It was adapted by Charles Grodin based upon the novel by Gerald A. Browne with the screenplay by Jeffrey Bloom.
  [Khalid_al-Habib.txt] Khalid al-Habib

Khalid al- Habib also believed to use the nom de guerre Khalid al Harbi is an Al- Qaeda leader who has overseen Al- Qaeda operations in Afghanistan. In 2005 it was reported that al- Habib died in a US missile strike on the Pakistani village of Damadola though it is unclear whether this is the case as Pakistani officials have claimed that no Al- Qaeda leaders died in the attack. al- Habib's nationality is not clear with sources claiming he may be Egyptian or Moroccan.
  ['Elisiva_Fusipala_Tauki'onetuku.txt] 'Elisiva Fusipala Tauki'onetuku

'Elisiva Fusipala Tauki'onetuku (18 May 1850 - September 1889) was the mother of King George Tupou II.

Step 1: apply_filter({
  "filter_type": "regex",
  "pattern": "Meek Mill",
  "target_set": "all",
  "output_prefix": "has_meek_mill"
})
  Agent thought: The performer of 'B Boy (song)' is Meek Mill, with guest appearances from Big Sean and ASAP Ferg. The question asks where the performer was detained. I will now search for information about Meek Mill being detained, focusing on location. I will start by searching for 'Meek Mill' in the corpus.
  Observation: {
  "positive_set": "has_meek_mill_pos",
  "positive_count": 2,
  "negative_set": "has_meek_mill_neg",
  "negative_count": 6116,
  "total_filtered": 6118,
  "sample_positive_docs": [
    {
      "doc_id": "B_Boy_song__638a840ec141",
      "filename": "B_Boy_song.txt"
    },
    {
      "doc_id": "Meek_Mill__52e01b469e2a",
      "filename": "Meek_Mill.txt"
    }
  ],
  "sample_negative_docs": [
    {
      "doc_id": "11_Harrowhouse__08cc6c95c2f9",
      "filename": "11_Harrowhouse.txt"
    },
    {
      "doc_id": "1971_Copa_Libertadores__8369d1b96cbe",
      "filename": "1971_Copa_Libertadores.txt"
    },
    {
      "doc_id": "Khalid_Abdel_Nasser__b9f61f804f4b",
      "filename": "Khalid_Abdel_Nasser.txt"
    },
    {
      "doc_id": "Khalid_al-Habib__ca209876aaad",
      "filename": "Khalid_al-Habib.txt"
    },
    {
      "doc_id": "\u021aerova__10c94b229856",
      "filename": "\u021aerova.txt"
    },
    {
      "doc_id": "\u02bbElisiva_Fusipala_Tauki\u02bbonetuku__3e99f95742e5",
      "filename": "\u02bbElisiva_Fusipala_Tauki\u02bbonetuku.txt"
    }
  ],
  "scope_check": "apply_filter searches DOCUMENT TEXT. 2 docs matched in text, but 0 filenames match the same pattern. If you intended to filter by filename/date, use search_filenames instead.",
  "keyword_decomposition": [
    {
      "keyword": "Meek",
      "matches": 3,
      "pct": 0.0
    },
    {
      "keyword": "Mill",
      "matches": 157,
      "pct": 2.6
    }
  ],
  "selectivity_note": "Combined pattern matched 2; individual keywords match 3\u2013157 each. Consider starting with one keyword.",
  "keyword_intersection": {
    "keywords": [
      "Meek",
      "Mill"
    ],
    "both_match": 2
  },
  "negative_entity_check": "279/6116 negatives contain 'Where' (keyword only)."
}
[Auto-sampled snippets (positive) -- 3 of 2 docs, explore further if needed]
  [B_Boy_song.txt] B Boy (song)

"B Boy" is a song by American hip hop recording artist Meek Mill. It was released as a third single from "Dreams Worth More Than Money" on January 3, 2015, by Maybach Music Group and Atlantic Records. The song, which was produced by Sap, features guest appearances from Big Sean and ASAP Ferg.
  [Meek_Mill.txt] Meek Mill

Robert Rihmeek Williams (born May 6, 1987), known professionally as Meek Mill, is an American rapper, songwriter, and activist. Born and raised in Philadelphia, he embarked on his music career as a battle rapper, and later formed a short-lived rap group, The Bloodhoundz. In 2008, Atlanta-based rapper T.I. signed Meek Mill to his first record deal. In February 2011, after leaving Grand Hustle Records, Mill signed with Miami-based rapper Rick Ross's Maybach Music Group (MMG). Mill's debut album, "Dreams and Nightmares", was released in 2012 under MMG and Warner Bros. Records. The album, preceded by the lead single "Amen" (featuring Drake), debuted at number two on the U.S. "Billboard" 200. In October 2012, Mill announced the launch of his own label imprint, Dream Chasers Records,...
[Auto-sampled snippets (negative) -- 3 of 6116 docs, explore further if needed]
  [11_Harrowhouse.txt] 11 Harrowhouse

11 Harrowhouse is a 1974 British DeLuxe Color film directed by Aram Avakian and starring Charles Grodin, Candice Bergen, James Mason, Trevor Howard and John Gielgud in Panavision. It was adapted by Charles Grodin based upon the novel by Gerald A. Browne with the screenplay by Jeffrey Bloom.
  [Khalid_al-Habib.txt] Khalid al-Habib

Khalid al- Habib also believed to use the nom de guerre Khalid al Harbi is an Al- Qaeda leader who has overseen Al- Qaeda operations in Afghanistan. In 2005 it was reported that al- Habib died in a US missile strike on the Pakistani village of Damadola though it is unclear whether this is the case as Pakistani officials have claimed that no Al- Qaeda leaders died in the attack. al- Habib's nationality is not clear with sources claiming he may be Egyptian or Moroccan.
  ['Elisiva_Fusipala_Tauki'onetuku.txt] 'Elisiva Fusipala Tauki'onetuku

'Elisiva Fusipala Tauki'onetuku (18 May 1850 - September 1889) was the mother of King George Tupou II.

=== Automatically computed ===
Filename search counts:
  search_filenames("Meek") -> 1 docs
  search_filenames("Mill") -> 12 docs
Set compositions -- has_b_boy_pos (1) x has_meek_mill_pos (2):
  intersect -> 1
  union     -> 2
  has_b_boy_pos \ has_meek_mill_pos -> 0
  has_meek_mill_pos \ has_b_boy_pos -> 1

Step 2: apply_filter({
  "filter_type": "regex",
  "pattern": "detain",
  "target_set": "has_meek_mill_pos",
  "output_prefix": "meek_mill_detain"
})
  Agent thought: The performer of 'B Boy (song)' is Meek Mill, as confirmed in the B_Boy_song.txt and Meek_Mill.txt documents. Now, I need to determine where Meek Mill was detained. I will search for the term 'detain' (to cover 'detained', 'detainment', etc.) in the Meek_Mill.txt document to find any mention of his detention and its location.
  Observation: {
  "positive_set": "meek_mill_detain_pos",
  "positive_count": 0,
  "negative_set": "meek_mill_detain_neg",
  "negative_count": 2,
  "total_filtered": 2,
  "sample_positive_docs": [],
  "sample_negative_docs": [
    {
      "doc_id": "B_Boy_song__638a840ec141",
      "filename": "B_Boy_song.txt"
    },
    {
      "doc_id": "Meek_Mill__52e01b469e2a",
      "filename": "Meek_Mill.txt"
    }
  ],
  "zero_result_hint": "Zero matches on 2 docs. Try: (1) a simpler/shorter regex, (2) search_filenames if the condition is about document names."
}

=== Automatically computed ===
Set compositions -- has_meek_mill_pos (2) x meek_mill_detain_pos (0):
  intersect -> 0
  union     -> 2
  has_meek_mill_pos \ meek_mill_detain_pos -> 2
  meek_mill_detain_pos \ has_meek_mill_pos -> 0

Step 3: apply_filter({
  "filter_type": "regex",
  "pattern": "arrest",
  "target_set": "has_meek_mill_pos",
  "output_prefix": "meek_mill_arrest"
})
  Agent thought: No matches for 'detain' or its variants in the Meek Mill-related documents. To ensure coverage, I will try searching for 'arrest' (a common synonym for 'detain') in the same set, as the corpus may use different terminology.
  Observation: {
  "positive_set": "meek_mill_arrest_pos",
  "positive_count": 0,
  "negative_set": "meek_mill_arrest_neg",
  "negative_count": 2,
  "total_filtered": 2,
  "sample_positive_docs": [],
  "sample_negative_docs": [
    {
      "doc_id": "B_Boy_song__638a840ec141",
      "filename": "B_Boy_song.txt"
    },
    {
      "doc_id": "Meek_Mill__52e01b469e2a",
      "filename": "Meek_Mill.txt"
    }
  ],
  "zero_result_hint": "Zero matches on 2 docs. Try: (1) a simpler/shorter regex, (2) search_filenames if the condition is about document names."
}

Step 4: apply_filter({
  "filter_type": "regex",
  "pattern": "prison|jail",
  "target_set": "has_meek_mill_pos",
  "output_prefix": "meek_mill_prison_jail"
})
  Agent thought: No matches for 'arrest' or 'detain' in the Meek Mill documents. To ensure I am not missing alternate terminology, I will search for 'prison' and 'jail' in the Meek Mill-related documents, as these are common terms used to describe where someone was detained.
  Observation: {
  "positive_set": "meek_mill_prison_jail_pos",
  "positive_count": 1,
  "negative_set": "meek_mill_prison_jail_neg",
  "negative_count": 1,
  "total_filtered": 2,
  "sample_positive_docs": [
    {
      "doc_id": "Meek_Mill__52e01b469e2a",
      "filename": "Meek_Mill.txt"
    }
  ],
  "sample_negative_docs": [
    {
      "doc_id": "B_Boy_song__638a840ec141",
      "filename": "B_Boy_song.txt"
    }
  ],
  "scope_check": "apply_filter searches DOCUMENT TEXT. 1 docs matched in text, but 0 filenames match the same pattern. If you intended to filter by filename/date, use search_filenames instead.",
  "keyword_decomposition": [
    {
      "keyword": "prison",
      "matches": 1,
      "pct": 50.0
    },
    {
      "keyword": "jail",
      "matches": 0,
      "pct": 0.0
    }
  ]
}
[Auto-sampled snippets (positive) -- 3 of 1 docs, explore further if needed]
  [Meek_Mill.txt] ...and "Ima Boss", being included on volume one. He released his second album, "Dreams Worth More Than Money", in 2015 and his third album, "Wins & Losses", in 2016. His fourth studio album, "Championships", was released in November 2018 and debuted at number one on the "Billboard" 200. Its lead single, "Going Bad" (featuring Drake), peaked at number 6 on the "Billboard" Hot 100. In November 2017, he was sentenced to two to four years in prison for violating parole and served five months at the State Correctional Institution - Chester in Chester, Pennsylvania, before being released while his trial continues. In August 2019, a documentary series about his battle with the criminal justice system, "Free Meek", was released on Amazon's Prime Video. Mill served as executive producer on the series.
[Auto-sampled snippets (negative) -- 3 of 1 docs, explore further if needed]
  [B_Boy_song.txt] B Boy (song)

"B Boy" is a song by American hip hop recording artist Meek Mill. It was released as a third single from "Dreams Worth More Than Money" on January 3, 2015, by Maybach Music Group and Atlantic Records. The song, which was produced by Sap, features guest appearances from Big Sean and ASAP Ferg.

=== Automatically computed ===
Filename search counts:
  search_filenames("prison") -> 3 docs
Set compositions -- meek_mill_arrest_pos (0) x meek_mill_prison_jail_pos (1):
  intersect -> 0
  union     -> 1
  meek_mill_arrest_pos \ meek_mill_prison_jail_pos -> 0
  meek_mill_prison_jail_pos \ meek_mill_arrest_pos -> 1
\end{lstlisting}

\section{Transcripts Dataset Construction}
\label{app:chrg}

This appendix describes the construction of the Transcripts corpus and question set used in our experiments (Section~\ref{sec:experiments}).  The goal is to produce questions where (i) all gold answers are \emph{derivable from the corpus text}, (ii) a substantial fraction of questions require \emph{corpus-wide aggregation} that a top-$k$ retrieval system structurally cannot answer, and (iii) the questions are phrased as a domain analyst would naturally phrase them, with no leading references to formatting features.

\paragraph{Source data.}
The corpus is drawn from the U.S.\ Government Publishing Office (GPO) collection of 117th-Congress hearing transcripts, distributed publicly via \texttt{govinfo.gov}.  We sample 75 transcripts, balanced across the House (39), the Senate (34), and joint House--Senate sessions (2), and spanning calendar 2021--2022.  Each transcript is converted to UTF-8 plain text from the GPO HTML release; document filenames retain GPO's public package-id convention: \texttt{CHRG-117hhrg\dots.txt} for House hearings, \texttt{CHRG-117shrg\dots.txt} for Senate hearings, and \texttt{CHRG-117jhrg\dots.txt} for joint House--Senate sessions. Chamber is therefore available in both the public metadata and each document's cover-page text.

\paragraph{Structural features used as gold.}
CHRG transcripts have a strongly conventional structure that makes deterministic gold construction tractable.  We rely on two families of features:
\begin{enumerate}\setlength{\itemsep}{2pt}
\item \textbf{Cover-page metadata.}  The first $\sim$160 lines of each transcript contain the chamber identifier (e.g., \texttt{U.S.\ HOUSE OF REPRESENTATIVES}, \texttt{UNITED STATES SENATE}, or a \texttt{JOINT HEARING} marker), the main committee heading (\texttt{COMMITTEE ON \dots}), an optional subcommittee heading (\texttt{SUBCOMMITTEE ON \dots}), the hearing date in the form \texttt{Weekday, Month Day, Year}, and the hearing title.  A parser extracts these four fields per document.%; a 75-document re-parse by an independent validator yields the same values for chamber, committee, and date on every transcript.
\item \textbf{Labeled-field markers in the body.}  CHRG transcripts use a small set of conventional line-anchored labels that recur across hearings: \texttt{Present:} and \texttt{Members present:} (formal attendance roll); \texttt{Staff Present:}; \texttt{Also present:} (visiting members); \texttt{Available via the World Wide Web:} (publication URL); the closing parliamentary notation \texttt{[Whereupon, at HH:MM \dots]}; and \texttt{Responses to written questions of \dots} (post-hearing QFR section header).  Each marker is detected by a line-anchored regex over the full transcript text; coverage counts (e.g., 50/75 transcripts contain an attendance roll under \texttt{Present:} $\cup$ \texttt{Members present:}; 26/75 adjourned during the 11 AM hour) are taken directly from these regex hits.
\end{enumerate}

\paragraph{Question types.}
All questions and gold answers are produced deterministically from the parsed corpus by a generation script (we release the script and the independent validator).  No hand-editing is applied.  The resulting evaluation file has 100 questions across four groups:
\begin{enumerate}\setlength{\itemsep}{2pt}
\item \textbf{Single-document lookups} (69/100): per-hearing questions about the committee, the chamber, the hearing date, and the number of witnesses listed in the Contents block.
\item \textbf{Chamber-level aggregations} (6/100): corpus-wide counts and comparisons across chambers, e.g., the total number of House versus Senate hearings, the size of the House--Senate difference, and which chamber predominates corpus-wide or within a year.
\item \textbf{Chamber $\times$ year intersections and year totals} (3/100): e.g., how many House hearings occurred in 2022; the total number of hearings in a given year.
\item \textbf{Labeled-field 3-way concept-majority questions} (22/100): given three corpus-level features (e.g., an attendance roll, a published online transcript link, and an 11~AM adjournment notation), the question asks which of the three appears in the most hearings; chamber-restricted variants ask the same within House- or Senate-only subsets.  %Each gold answer is the feature with the strictly largest hit count, and we keep only triples whose magnitude gap is robust to broader natural-language readings of the concept.
\end{enumerate}

\paragraph{Natural-language phrasing without leading hints.}
Both the question text and the gold answers are phrased in natural language, without quoted regex markers or formatting hints.
%Concretely, the generator refers to each labeled-field concept by a descriptive natural-language phrase (e.g., \emph{an attendance line at the start of the hearing showing the committee members physically present}, rather than ``a `Present:' line''), and avoids mentioning the cover page or the literal \texttt{COMMITTEE ON \dots} heading in the question text.  The motivation is twofold: (i) the question should be answerable by reading the transcript as a human would, not by pattern-matching a marker the generator chose; and (ii) such phrasing eliminates the trivial path where an agent simply pastes the quoted marker into a regex. 
The two semantically equivalent attendance markers (\texttt{Present:} and \texttt{Members present:}) are merged into a single concept so that a natural reading of the question is unambiguous.

\paragraph{Independent validation of all golds.}
Every gold answer is recomputed from the corpus by an independent validator
using a separate code path from the generator. It re-derives each count,
intersection, and three-way majority from scratch; the two implementations
agree on all 100 questions. We additionally inspect five randomly selected
documents to confirm that matches align with the natural reading of the text.

% \paragraph{Answer formatting and unambiguity.}
% Each question has exactly one correct answer.  Gold answers match the minimal format implied by the question: count questions $\rightarrow$ a single integer; majority questions $\rightarrow$ the short label of the winning feature; single-document lookup questions $\rightarrow$ the literal value as it appears in the transcript (the committee heading, the formatted date, or the chamber name).  Question types whose gold is not robustly recoverable from the corpus under a reasonable natural reading were excluded during generator design.

\section{LLM-as-a-Judge Details}
\label{app:judge}

We evaluate all systems using the same LLM-as-a-judge pipeline to avoid system-specific evaluators. The judge returns a score in $[0,1]$ and may assign fractional credit under the rubric below; tables report the mean score across questions multiplied by 100. We therefore call this quantity \emph{judge score}, rather than binary accuracy. For each question, we run two judge-model calls:
\begin{enumerate}\setlength{\itemsep}{2pt}
\item \textbf{Answer-only extraction:} transform a possibly verbose model output into a single ``final answer'' string (or \texttt{INSUFFICIENT\_EVIDENCE}).
\item \textbf{Judging:} compare the extracted final answer against the gold answer and return a JSON verdict and score.
\end{enumerate}

Prompt for answer-only extraction (reasoning removal):

\begin{lstlisting}
You are a QA assistant.

You will be given:
- the question
- a model answer that may contain reasoning

Your task:
- Output ONLY the final answer to the question.
- Do NOT include any reasoning, explanation, preamble, or extra text.
- Do NOT use markdown code fences.

If the answer cannot be determined from the provided model answer, output exactly:
INSUFFICIENT_EVIDENCE
\end{lstlisting}

SQuAD-style exact-match and token-overlap F1 are computed on the output of the above normalization.

Prompt for judging rules and scoring:

\begin{lstlisting}
You are an accurate evaluator for QA.
You will be given:
- the question
- the gold answer
- the model answer (final answer only; no reasoning)

Decide if the model answer matches the gold answer.
Be strict about factual correctness, but allow paraphrases and equivalent numeric formats.
If the gold answer is empty, mark as incorrect unless the model explicitly says it cannot be determined (e.g., INSUFFICIENT_EVIDENCE).

INSUFFICIENT_EVIDENCE rule:
- If the model answer is exactly the string "INSUFFICIENT_EVIDENCE" AND the gold answer is non-empty, verdict MUST be "incorrect" and score MUST be 0.

Special scoring rules:

1) Count questions:
- If the question is asking for a count/number, the answer is ONLY correct if the predicted number exactly matches the gold number.
- If it is off by even 1, score MUST be 0 and verdict MUST be "incorrect".

2) List/set questions:
- Treat the gold answer and model answer as sets of items.
- Let L be the number of distinct gold items.
- Start with score=1.0.
- Apply a penalty of (1/L) for each missing gold item AND for each extra predicted item.
- Treat possible aliases as correct. Do not give partial score per element.
- Score is floored at 0.
- If the model includes additional explanatory text, ignore it and focus on the item set.

When the answer is a proper name / named entity:
- Treat clear aliases or alternative spellings as correct (score=1).
- If the model answer is ambiguous (could refer to multiple entities) but not clearly wrong, allow partial credit with a short rationale.
- Give score=0 only if it is clearly the wrong entity.

Return a JSON object with:
- verdict: one of ["correct","incorrect","partial"]
- score: number in [0,1] where correct=1, incorrect=0, partial is between
- rationale: short explanation (1-3 sentences)
- abs_diff: OPTIONAL. Include ONLY for count/number questions. It must be a non-negative integer equal to |gold_number - predicted_number| if you can extract both numbers; otherwise null.

Rules:
- Output must be valid JSON (no markdown fences, no extra text).
\end{lstlisting}

\end{document}